\documentclass[sigconf]{acmart}

\AtBeginDocument{%
  \providecommand\BibTeX{{%
    \normalfont B\kern-0.5em{\scshape i\kern-0.25em b}\kern-0.8em\TeX}}}

\setcopyright{acmcopyright}
\copyrightyear{2018}
\acmYear{2018}
\acmDOI{10.1145/1122445.1122456}

\acmConference[Woodstock '18]{Woodstock '18: ACM Symposium on Neural
  Gaze Detection}{June 03--05, 2018}{Woodstock, NY}
\acmBooktitle{Woodstock '18: ACM Symposium on Neural Gaze Detection,
  June 03--05, 2018, Woodstock, NY}
\acmPrice{15.00}
\acmISBN{978-1-4503-XXXX-X/18/06}

\usepackage{amsmath}
\usepackage{multirow}
\usepackage{bbm}
\usepackage{ulem}
\usepackage{graphicx}
\usepackage{subfigure}
\usepackage{algorithm}
\usepackage{url}
\usepackage{algorithmic}
\makeatletter
\newcommand{\rmnum}[1]{\romannumeral #1}
\newcommand{\Rmnum}[1]{\expandafter\@slowromancap\romannumeral #1@}
\newcommand{\vectorproj}[2][]{\textit{proj}_{\vect{#1}}\vect{#2}}
\newcommand{\vect}{\mathbf}
\makeatother
\begin{document}

%\title{Semi-supervised Learning from Stream Data with Evolving Distribution Shift}
%\title{Integrated Robust Labeling and Effective Adaption for  Semi-supervised Drifted Stream Learning}
\title{Integrated Robust Labeling and Anti-forgetting Adaptation for  Semisupervised Drifted Stream Learning with Short Lookback}
\title{Semi-supervised Drifted Stream Learning with Short Lookback}

\author{Weijieying Ren}
\affiliation{%
  \institution{University of Central Florida}\country{USA}}
\email{wjyren@knights.ucf.edu}

\author{Pengyang Wang}
\affiliation{%
  \institution{University of Macau}
  \country{China}}
\email{pywang@um.edu.mo}

\author{Xiaolin Li}
\affiliation{%
  \institution{Nanjing University}
  \country{China}}
\email{lixl@nju.edu.cn}

\author{Charles E. Hughes}
\affiliation{%
  \institution{University of Central Florida}
  \country{USA}}
\email{charles.hughes@ucf.edu}

\author{Yanjie Fu}
\affiliation{%
  \institution{University of Central Florida}
  \country{USA}}
\email{yanjie.fu@ucf.edu}


\vspace{-0.3cm}
\begin{abstract}
In many scenarios, 1) data streams are generated in real time; 2) labeled data are expensive and only limited labels are available in the beginning;  3) real-world data is not always i.i.d. and data drift over time gradually; 4) the storage of historical streams is limited and model updating can only be achieved based on a very short lookback window.
This learning setting limits the applicability and availability of many Machine Learning (ML) algorithms. 
We generalize the learning task under such setting as a semi-supervised drifted stream learning with short lookback problem (SDSL).
SDSL imposes two under-addressed challenges on existing methods in semi-supervised learning, continuous learning, and domain adaptation: 1) robust pseudo-labeling under gradual shifts and 2) anti-forgetting adaptation with short lookback.
To tackle these challenges, we propose a principled and generic generation-replay framework to solve SDSL. 
The framework is able to accomplish: 1) robust pseudo-labeling in the generation step; 2) anti-forgetting adaption in the replay step. 
To achieve robust pseudo-labeling, we develop a novel pseudo-label classification model to leverage supervised knowledge of previously labeled data, unsupervised knowledge of new data, and, structure knowledge of invariant label semantics. 
To achieve adaptive anti-forgetting model replay, we propose to view the anti-forgetting adaptation task as a flat region search problem. We propose a novel minimax game-based replay objective function to solve the flat region search problem and develop an effective optimization solver. Finally, we present extensive experiments to demonstrate our framework can effectively address the task of anti-forgetting learning in drifted streams with short lookback.

%Semi-supervised learning, i.e. jointly learning from labeled and unlabeled samples, is an active research topic due to its key role on relaxing annotation burden. Current SSL methods have made substantial advance in dealing with I.I.D and static unlabeled data. However, this can be unrealistic in real-world applications, where unlabeled data usually come in an online and continually evolving manner, pose challenges to classic SSL paradigm: (1) Mainstream SSL methods are tailored to stationary data, and can fail in non-stationary environments.  (2) Since data evolves continually, poses a great challenge to generate high-quality pseudo labels. (3) When the unlabeled data arrive online, the model should also maintain competence on previous learned data, i.e. adapt without forgetting. Storing previous data provides one way to remember learned knowledge, while such replay strategy throws a new challenge to resource restrictions. However, the work on this field is rather limited. Therefore, we study a novel problem of semi-supervised learning via gradual shifting environment. Our framework consists of two components: a target task oriented classifier which generate high-quality pseudo labels by mining unlabeled data structure as well as preserve class-wise invariant relations; and rectified retraining objective which adapts on new shifted data without forgetting, reserving knowledge from previous learned data. Extensive experiments show that our approach achieves significant improvement over related state-of-the-art methods.
\end{abstract}

\settopmatter{printfolios=true}
\maketitle
\vspace{-0.3cm}
\section{Introduction}
\begin{figure}[t]
\vspace{-0.0cm}
  \centering
  \includegraphics[width=\linewidth]{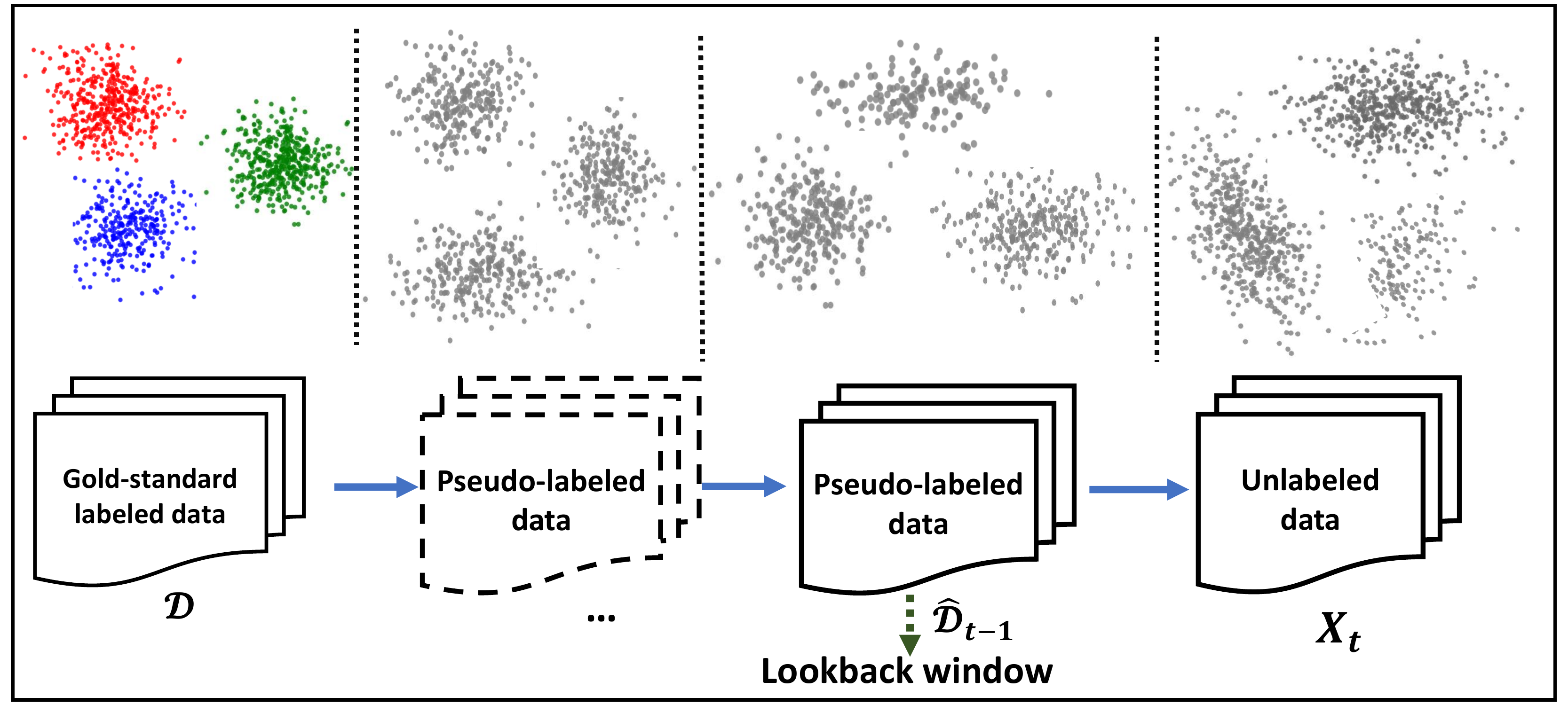}
  \vspace{-0.3cm}
\caption{Learning in the semi-supervised, stream, gradually drifted data environment with short lookback.}
\vspace{-0.5cm}
\label{fig:train_setting}
\end{figure}

Considering a motivating application of in-App activity analysis. Many mobile Apps, such as Snapchat, generate unlabeled internet traffic streams in real time. In-App activities (e.g., share photos, videos, text, and drawings) could drift over time, resulting in distribution shifts. 
Due to mobile privacy concerns, many companies implement a very short data retention duration policy. 
We only have access to the most recent data (e.g., a lookback window). 
Therefore, learning with short lookback in unlabeled drifted streams is critical for in-App activity classification. 
This scenario can be generalized as a new learning problem: \textit{\textbf{S}emi-supervised \textbf{D}rifted \textbf{S}tream learning with short \textbf{L}ookback (SDSL)}. which is depicted in Figure \ref{fig:train_setting}.
SDSL can enable a model to adaptively learn from an unlabeled stream of evolving distribution shifts, with limited initial labels and small lookback windows for training.
Solving SDSL can address multiple critical issues to increase the availability and applicability of ML algorithms.
For example, in many scenarios, 1) data are generated in real time; 2) labeled data are expensive and only limited labels are available in the beginning of time; 3) real-world data is not always i.i.d. and data drift over time gradually; 4) the storage of historical streams is limited and model updating can only be achieved based on a very short lookback window. 

There are two major challenges in solving SDSL: 1) robust pseudo-labeling under distribution shifts, 2) anti-forgetting adaptation with short lookback. 
Firstly, except for the initially given labels, all the incoming streams are unlabeled. 
The evolving distribution shifts further introduce bias into the task of labeling new data: training a classifier on the previously labeled data, but utilizing the classifier to predict labels of forthcoming drifted data. 
Robust pseudo-labeling is to answer: how can we generate robust and quality pseudo-labels of unlabeled streams to augment old data?
Secondly, SDSL can suffer from forgetting. This is because:
1) due to stream storage limitations and privacy-driven short data retention policies, less old data are stored for retraining;
2) when a model adapts to new drifted data, the model parameters change to fit new data and forget old knowledge.  
A model that forgets old knowledge will perform poorly when new data with old distribution re-appear in the future stream. 
Therefore, anti-forgetting adaptation under short lookback is to answer: how can we learn new knowledge, while prevent forgetting old knowledge with limited historical data for replay?

Relevant works can only partially solve SDSL.
Firstly, SDSL is related to Semi-Supervised Learning (SSL) algorithms~\cite{guo2020record,berthelot2019mixmatch}, which combines a small amount of labeled data with a large amount of unlabeled data during training. 
However, in classic SSL, 1) the labeled and unlabeled data are assumed to sample from an i.i.d distribution; 2) unlabeled data is static without evolving shift over streams~\cite{berthelot2019mixmatch}. 
Secondly, SDSL is related to continual learning that learns knowledge of new data without forgetting learned knowledge of old data. 
However, classic continual learning assumes newly generated data in streams are all labeled~\cite{pan2010domain,rolnick2019experience}.
Thirdly, SDSL is related to domain adaptation ~\cite{tzeng2017adversarial,zhao2020semi} that aims to transfer knowledge learned from one or multiple labeled source domains to an unlabeled target domain.
However, in classic domain adaption, both source domains and target domains are static~\cite{pan2010domain,rolnick2019experience}. 
Existing studies demonstrate the inability to jointly address both robust pseudo-labeling and anti-forgetting adaptation with short lookback in SDSL. 
As a result, it highly necessitates a novel perspective to derive the novel formulation and solver of SDSL. 

\textbf{\textbf{Our Contribution: an integrated robust and antiforgetting perspective.}}
We formulate a generic learning problem of SDSL for semi-supervised, stream, limited lookback memory, evolving distribution shift environments. 
We show that semi-supervised learning in streams can be solved by iterating the label generation and the model replay process, where the label generation is to generate pseudo-labels for newly coming unlabeled data stream, and the model replay is to retrain learning model with old data of short lookback and pseudo-labeled new data. 
Robust pseudo-labels are important for effective SDSL. 
We find that leveraging invariant structure knowledge in streaming data can fight against the bias introduced by evolving distribution drifts for the robust pseudo-label generation. 
We demonstrate that to better learn patterns at the transition between old and new data, the model replay needs to be adaptive while anti-forgetting, even with a short lookback window.
This requirement can be reformulated into a task of automated identification of flat regions.
We highlight that the automated flat gradient region identification problem is indeed a minimax game.
Solving the minimax game can effectively help the model to achieve both the anti-forgetting replay with limited old lookback data and adaption to new data.

\textbf{Summary of Proposed Approach}. Inspired by these findings, this paper presents the first attempt to develop a principled and generic generation-replay framework for the SDSL problem by iterating the robust pseudo-label generation and the adaptive anti-forgetting model replay.
The framework has two goals: 1) robust pseudo-label generation against evolving distribution shifts in the generation step; 2) balancing anti-forgetting replay and effective adaptation in the replay step. 
%
%Specifically, to achieve the Goal 1, we argue that the unrobustness of existing pseudo label generation methods is caused by training a classifier on previous labeled data, but utilizing the classifier to predict labels of forthcoming drifted data.  The temporal drift leads to the inability of  the trained classifier to predict robust labels for unseen drifted data. 
Especially, to achieve Goal 1, we develop a three-step robust label generation method to leverage multi-level knowledge.
In particular, we find that robustness of pseudo-labels can be improved by modeling supervised knowledge from previously labeled data, unsupervised knowledge from new unlabeled drifted data, and structure knowledge from invariant label class semantics. 
Step 1 develops a supervised neural encoder-based classifier trained on previously labeled data. Then, We adopt a center-based clustering method to adjust labels on new drifted data, and leverage the invariance of label class semantics to regularize the neural encoder-based classifier. 
To achieve Goal 2, we develop an adaptive anti-forgetting model replay technique. 
In particular, we reformulate the adaptive anti-forgetting model replay into a flat region search problem. We propose a novel minimax game-based replay objective to automatically find the flat region to  minimize the predictive loss on both previous data and new drifted data. 
And we develop an effective optimization method to solve the minimax game.
Finally, we present extensive empirical results to demonstrate the effectiveness of our method for learning in the semi-supervised, streaming, gradually drifted, and short lookback setting.

\vspace{-0.3cm}
\section{Problem Statement}
\textbf{The SDSL Problem.} 
Let use $\mathcal{D}$ to denote gold standard label data at $t$=0  and use $\hat{\mathcal{D}}$ to denote pseudo-labeled data at $t$>0.
Considering the existence of the initial labeled data that includes a feature matrix $\mathbf{X}_0$ and a gold standard label vector $\mathbf{y}_0$ in the very beginning (denoted by  $\mathcal{D}=\{\mathbf{X}_0, \mathbf{y}_0\}$ ), and a drifted unlabeled data stream (denoted by a feature matrix list of stream segmentations $ [\mathbf{X}_t]_{t=1}^{T}$). 
We aim to train an adaptive model to classify all the data points of the unlabeled data stream into a fixed number of classes. 
At the time $t$, the model generates a list of pseudo-labeled sets $[\mathbf{y}_1,..., \mathbf{y}_{t-1}]$ for the list of unlabeled stream segmentations $[\mathbf{X}_1,..., \mathbf{X}_{t-1}]$.

\begin{figure}[t]
  \centering
  \includegraphics[width=\linewidth]{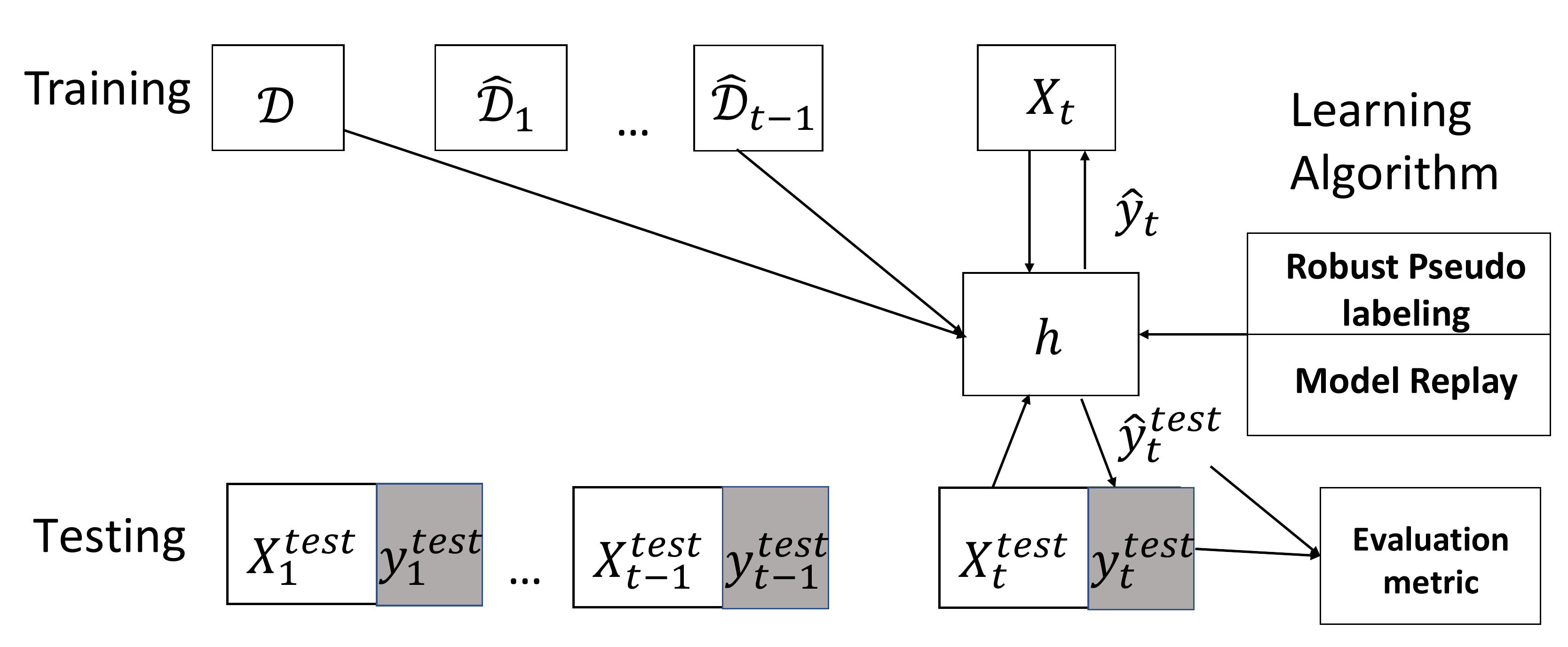}
\vspace{-0.5cm}
\caption{The training and testing stage of SDSL.}
\label{fig:train}
 \vspace{-0.5cm}
\end{figure}
During the training phase (Figure \ref{fig:train}), we iteratively learn the adaptive model $h$ that takes only the initial gold standard labeled data $\mathcal{D}$ and the short lookback of the ($t$-1)-th pseudo-labeled stream segmentation (denoted by $\hat{\mathcal{D}}_{t-1} = \{X_{t-1}, \hat{\mathbf{y}}_{t-1}\}$) as inputs, and predicts the pseudo-labels (denoted by $\hat{\mathbf{y}}_t$) of the $t$-th stream segmentation (denoted by $\mathbf{X}_{t}$) at the time $t$.
Formally, the approximation function $h$ with learning parameters $\theta$ is given by:
\begin{equation}
    h_{\theta}(\mathcal{D}, \hat{\mathcal{D}}_{t-1}, \mathbf{X}_{t}) \rightarrow \mathbf{\hat{\mathbf{y}}}_t.
\end{equation}
The optimization objective is to learn the model that can 
1) generate robust pseudo-labels,
%to \textbf{augment unlabeled stream data} 
and 2) prevent forgetting old data while adapt well to new data. 
\vspace{-0.1cm}
\section{The Generation-Replay Framework}

\subsection{Overview}
\begin{figure}[t]
\vspace{-0.3cm}
  \centering
  \includegraphics[width=\linewidth]{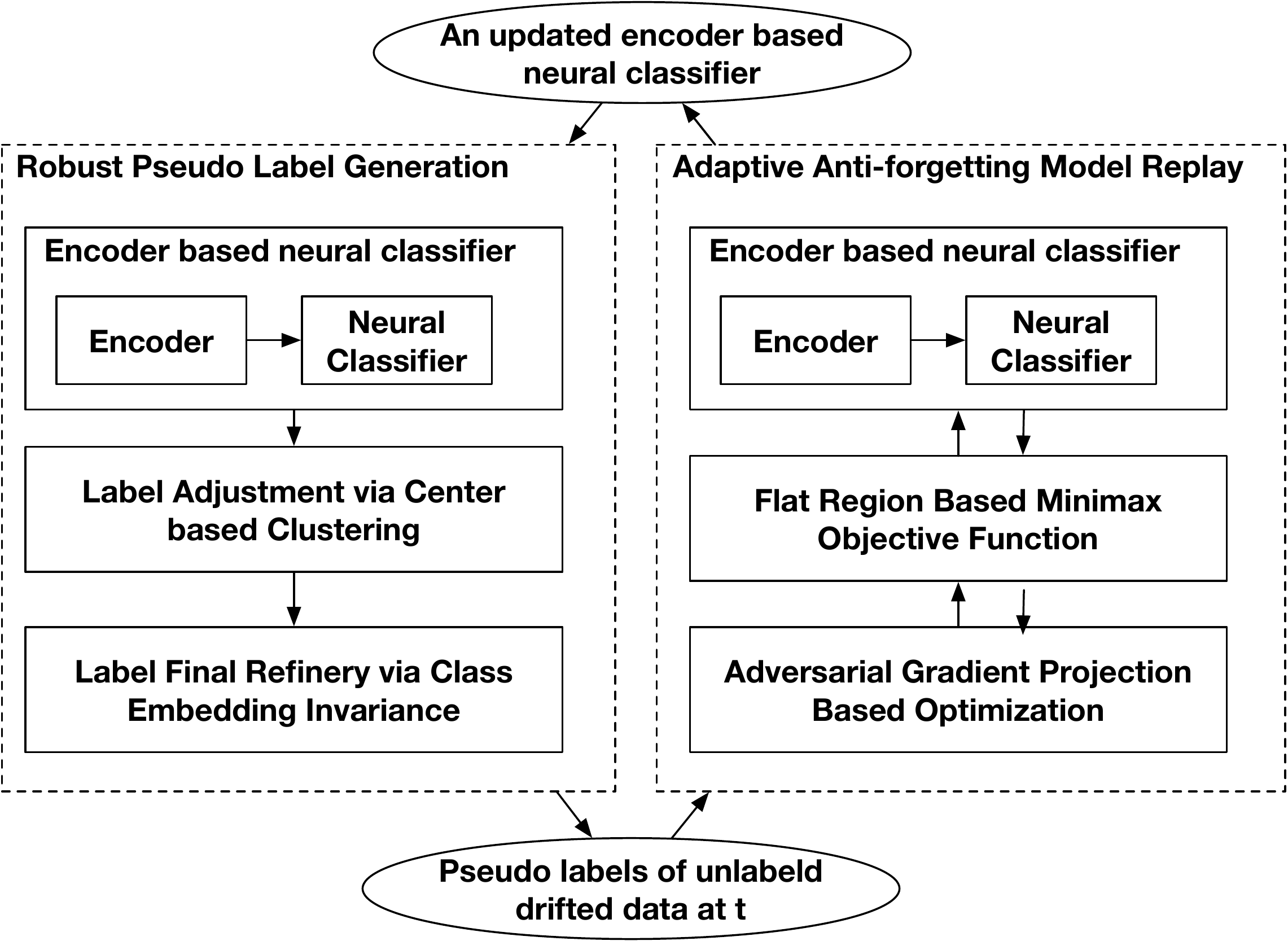}
 \vspace{-0.7cm}
\caption{Overview of the proposed framework. }
\label{fig:overview}
\vspace{-0.6cm}
\end{figure}

Figure~\ref{fig:overview} shows our proposed generation-replay framework including two components: 
1) pseudo label generation; 
2) adaptive anti-forgetting model replay. 
To achieve robust pseudo-label generation, 
we propose a three-step approach to improve the robustness of generated pseudo-labels by leveraging the knowledge from previously labeled data, new unlabeled data, and invariant label classes semantics, i.e., class correlations.
In particular, in Step 1, we first train an integrated encoder and a classifier by minimizing the predictive loss on previous gold-labeled data and pseudo-labeled data at $t$-1.  
We then use the trained classifier to predict and obtain the initial pseudo-labels of forthcoming unlabeled drifted data at $t$. 
Since the trained neural classifier learns previous knowledge that is partially overlapped with the knowledge of forthcoming drifted data, the labels of new drifted data that are non-overlapped with previous knowledge could be biased and inaccurate. 
In Step 2, we then propose to use an unsupervised centroid-based clustering method to adjust the labels of new drifted data by repeatably assigning the new drifted data to the nearest class centroid and updating class centroids until converged. 
The adjusted pseudo-labels of the forthcoming unlabeled drifted data are compared with model predicted labels to create loss signals as feedback to improve the neural integrated encoder-classifier method. 
Besides, we find that the semantic meanings of label classes remain invariant during the SDSL. 
We propose to leverage the invariance property of label class embeddings to further refine the class centroids. Specifically, we optimize the label class centroid via fixing label latent embedding learned from initial gold-labeled data.  %optimizing the label class centroid matrix factorization task while 
After obtaining the refined class centroids, we reassign forthcoming drifted data to the nearest class cluster centroid to generate debiased and robust labels.  
To achieve adaptive anti-forgetting model replay, we aim to retrain the integrated encoder-classifier to prevent forgetting previous knowledge while adapting well to new drifted data. 
We formulate this joint objective into a problem of searching the flat region.
It is challenging to not just search the flat region but also identify the width of the flat region. 
We found that the challenge can be converted into a formulation of solving a minimax game. 
We develop an effective optimization method to solve the minimax game to find the flat region and identify its width. 

\subsection{Robust \textbf{Pseudo Label} Generation}
\textbf{Why Robust \textbf{Pseudo Label} Generation Matters.} Conventional semi-supervised learning generates pseudo-labels based on the predictions of a supervised model on labeled data (i.e., pseudo-label generation stage) and then integrates the pseudo-labeled new data to retrain an updated model (i.e., replay stage). In this way, the network gradually generalizes to unlabeled data in a self-paced curriculum learning manner \cite{lee2013pseudo}. 
However, such a strategy is not applicable to the SDSL setting because the unlabeled stream data drift over time. 
Be sure to notice that, the supervised model is trained based on previously labeled data. 
When the supervised model predicts on forthcoming drifted unlabeled data, the generated pseudo-labels are likely to be biased and inaccurate, which will propagate errors to the replay stage.

\begin{figure}[t]
\vspace{-0.3cm}
  \centering
  \includegraphics[width=\linewidth]{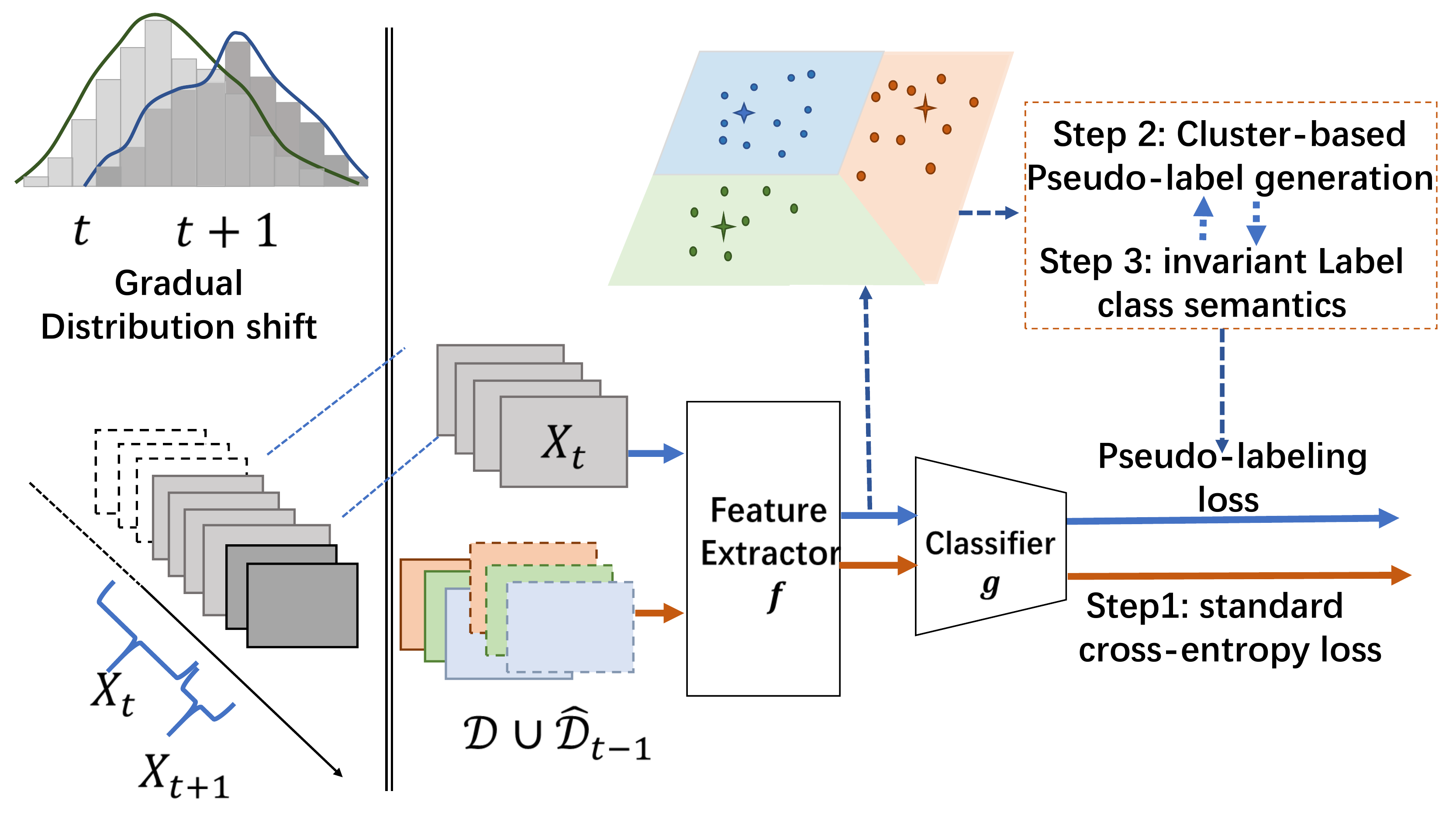}
 \vspace{-0.7cm}
\caption{The Robust Pseudo-Label Generation Process. }
\label{fig:label_relation}
\vspace{-0.5cm}
\end{figure}

\noindent\textbf{Leveraging Multi-level Knowledge to Robustify Pseudo-Label Generation.} 
We find that label generation of unlabeled drifted data under SDSL can be robustized by integrating supervised knowledge from previously labeled data, unsupervised knowledge from new unlabeled drifted data, and structure knowledge from invariant label class semantic meanings and relationships.
Based on our unique insight, we propose a step-by-step testable method that includes three steps. Figure \ref{fig:label_relation} illustrates the framework of the robust pseudo-label generation process.

\noindent\textbf{ Step 1: Leveraging Supervised Knowledge}
In a stream, data distributions drift gradually. In other words, the distribution of forthcoming unlabeled drifted data partially overlaps with the distribution of previous data.  
As shown in Figure \ref{fig:label_relation}, historical gold-labeled data and pseudo-labeled data still contain useful knowledge that can be used to predict the overlapped part of forthcoming drifted unlabeled data.

To this end, we develop an encoder-based neural classification model. This model jointly includes a multi-layer neural encoder $f$ and a neural classifier $g$ as an approximation function of the  classification model: $h = g(f(\cdot))$. 
The neural encoder takes a data point as input and outputs the embedding of the data point. The neural classifier takes the embedding of the data point as features and outputs the predicted labels. 
Aside from the approximation function, we use the \textbf{C}ross-\textbf{E}ntropy (CE) loss $\ell(\cdot, \cdot)$ to measure classification errors. Formally, the optimization objective function can be formulated as:
\begin{equation}
\label{classifier}
  \min_{\theta}\mathcal{L}_{CE}(\mathcal{D}, \hat{\mathcal{D}}_{t-1};\theta) = \sum_{(\mathbf{x}_i, y_i) \in \{\mathcal{D} \cup \mathbf{\hat{\mathcal{D}}}_{t-1}\}} \ell(h_\theta(\mathbf{x}_i),y_i).
\end{equation}
%Finally, we minimize the cross-entropy loss to learn the parameters of the encoder-based neural classifier. 
In this way, we leverage the gold labels and previous pseudo-labels of historical data as supervision signals to learn the encoder-based neural classifier to further generate initial pseudo-labels for current unlabeled data $\mathbf{X}_t$, which will be introduced next.

\noindent\textbf{ Step 2: Leveraging Unsupervised Knowledge.}
Since the trained encoder-based neural classifier learns previous knowledge that is partially overlapped with the knowledge of forthcoming drifted data, the labels of new drifted data that are non-overlapped with previous data distributions could be biased and inaccurate. How can we improve the label quality of new drifted data whose patterns are not seen in the knowledge of previous data?

We find that unsupervised knowledge in new drifted unlabeled data is helpful for improving and refining the labels of such data themselves. 
Centroid-based clustering is an unsupervised learning method to exploit unsupervised information to discover data grouping patterns.
Different from using the centroid-based clustering for data grouping, we propose to use such a method for data label adjustment. 
The high-level idea is to exploit the centroid-based clustering to adjust the labels of new drifted data by repeatably assigning the new drifted data to the nearest class centroid and updating class centroids until converged. 
The underlying insight is that label class centroid-based clustering reassigns labels based on the global pattern structure of new unlabeled drifted data. 
%\textbf{Such label adjustment is less sensitive to biased or wrong labels generated by Step 1. }

Specifically, in Step 2, we firstly use the trained encoder-based neural classifier to classify the labels of new drifted unlabeled data. We then use these classified labels to compute the centroids of all the label classes. 
In the initialization of the centroid-based clustering, we exploit all the class centroids as the initial cluster centroids. Formally, 
the centroid embedding in class $c$ are initialized as via:
\begin{equation}
\label{eq:initialization}
\mathbf{u}_c^{(0)} = \frac{\sum_{\mathbf{x}_i \in \mathbf{X}_t} \delta (h(\mathbf{x}_i))f(\mathbf{x}_i)}{\sum_{\mathbf{x}_i \in \mathbf{X}_t}\delta (h(\mathbf{x}_i))},
\end{equation}
where $\delta(\cdot)$ is the softmax function \cite{liu2016large}.
%with $\delta(\mathbf{a}_k) = \frac{exp(a_k)}{\sum_i^K(exp(a_i))}$ denoting the $k$-th element in the softmax output of a $K$ dimensional vector $\mathbf{a}$.  
We then repeat two tasks:  assigning data points to the nearest class and updating class centroids, until converged (e.g., maximum number of iterations). 
Particularly, in the data reassignment task, given the class centroid $\mathbf{u}_k$, we construct the nearest centroid classifier to assign each unlabeled data point $x_i \in \mathbf{X}_t$ to a class cluster as:
\begin{equation}
\label{eq:label}
    \hat{y}_i = \arg \min_{c \in C} d(f(\mathbf{x}_i), \mathbf{u}_{c}),
\end{equation}
where $d(\cdot, \cdot)$ measures the cosine distance between data $\mathbf{x}_i$ and the class centroid $\mathbf{u}_c$. And $C$ denotes class numbers.
%In the centroid updating task, 
Given the new pseudo-labels, we update the class centroids at each iteration $k$ by:
\begin{equation}
\label{eq:centroid}
    \mathbf{u}_c^{(k)} = \frac{\sum_{\mathbf{x}_i \in \mathbf{X}_t}f(\mathbf{x}_i) * \mathbbm{1} (y_i ==1)}{\sum_{\mathbf{x}_t \in \mathbf{X}_i} \mathbbm{1} (y_{i} ==c)},
\end{equation}
where $\mathbbm{1}(\cdot)$ is the indicator function. 
The centroid-based clustering can reduce the error caused by supervised prediction under drifted data, and generate adjusted pseudo-labels. 

Finally, the adjusted pseudo-labels of the forthcoming unlabeled drifted data are compared with model predictions to create loss signals as feedback to improve the training of the encoder-based neural classifier. 
The objective function can be formulated as:
\begin{equation}
\label{classifier}
  \min_{\theta}\mathcal{L}_{PL}(\mathbf{X_t;\theta})= \sum_{\mathbf{x}_i \in \mathbf{X}_t} \ell(h_\theta(x_i),
  \hat{y}_i).
\end{equation}
\noindent\textbf{ Step 3: Leveraging Structure Knowledge of Invariant Label semantics}
%Label Class Embedding Invariance.}
Intuitively, the semantic meanings of label classes remain invariant during the SDSL. 
We show that leveraging the invariance property of label class semantics can further refine the quality of the generated pseudo-labels of new drifted data.  
Our insight is based on the invariance property of label classes. The embeddings of label classes should remain invariant over timelines.
%from $t$=0 to $T$.
%To exploit such regularization to improve the quality of pseudo-labels, we propose to use the matrix factorization method \cite{} of label class centroid matrix to 
%bridge the gap between label reassignment and label class embedding. 

The underlying idea is that label quality can be improved by reassigning data points based on improved label class centroids.
To improve the accuracy of label class centroids, we can treat label class centroids as a  matrix and exploit factorization-based matrix reconstruction to reconstruct improved label class centroids. 
%Be sure to notice that, 
Notablely, factorizing the class centroid matrix can factorized into the embeddings of features and the embeddings of label classes, which links to the invariance regularization of label class embedding.

Specifically, in Step 3, we first obtain the gold standard label class embedding $\hat{\mathbf{V}}$ by factorizing the class centroid matrix $\mathbf{U}$ of the gold standard labeled data ($\mathcal{D}$) at $t$=0. We then perform factorization-based class centroid matrix reconstruction, by fixing the embedding of label classes $\hat{\mathbf{V}}$ to that learned from initial gold standard label data at $t$=0. 
Formally, the regularization term is defined as:
\begin{equation}
\mathcal{R}(\mathbf{U}_t) = \min_{\mathbf{U}_t,\mathbf{H}_t} 
\|\mathbf{U}_{t} - \mathbf{H}_{t}^{T} {\hat{\mathbf{V}}}\|_2.\\
 \label{Eq:semantics}
\end{equation}
With the given label class embedding $\hat{\mathbf{V}}$, we update class centroids $\mathbf{U}_t$ and feature latent embeddings $\mathbf{H}_{t}$ iteratively.
By solving the optimization problem, we obtain the refined class centroids. Finally, we reassign forthcoming drifted data to the nearest class cluster centroid to generate debiased and robust pseudo-labels. 

\textbf{Final loss function} The final objective function of robust pseudo-label generation can be represented as :
\begin{equation}
\label{classifier}
\mathcal{L}_{total} =\mathcal{L}_{CE}(\mathcal{D}, \mathcal{\hat{D}}_{t-1};\theta) + \mathcal{L}_{PL}(\mathbf{X}_t;\theta) + \mathcal{R}(\mathbf{U}_t),
\end{equation}
where $\mathcal{L}_{CE}(\mathcal{D}, \mathcal{\hat{D}}_{t-1};\theta)$ represents cross-entropy loss on gold-labeled data $\mathcal{D}$ and pseudo-labeled data $\mathcal{\hat{D}}_{t-1}$. Besides, $\mathcal{R}(\mathbf{U}_t)$ stabilize the updating of centroids which contains invariant label class semantics. %regularizes the centroids updating.
A detailed optimization process can be referred to Algorithm 1 in Appendix A.1.

\subsection{Adaptive Anti-forgetting Model Replay}
After generating robust pseudo-labels for the newly coming unlabeled drifted data at time $t$, we treat such pseudo-labeled data as part of training data, and retrain the encoder-based neural classifier using the gold label data $\mathcal{D}$ and the newly generated pseudo-labeled data $\hat{\mathcal{D}_t}$ at time $t$. To simplify our description, we ignore $\mathcal{D}$ in the following sections, since $\mathcal{D}$ is available across timelines.

\noindent\textbf{Why Anti-forgetting Adaptation Matters?} 
\begin{figure}[t]
  \centering
  \includegraphics[width=\linewidth,height = 4cm]{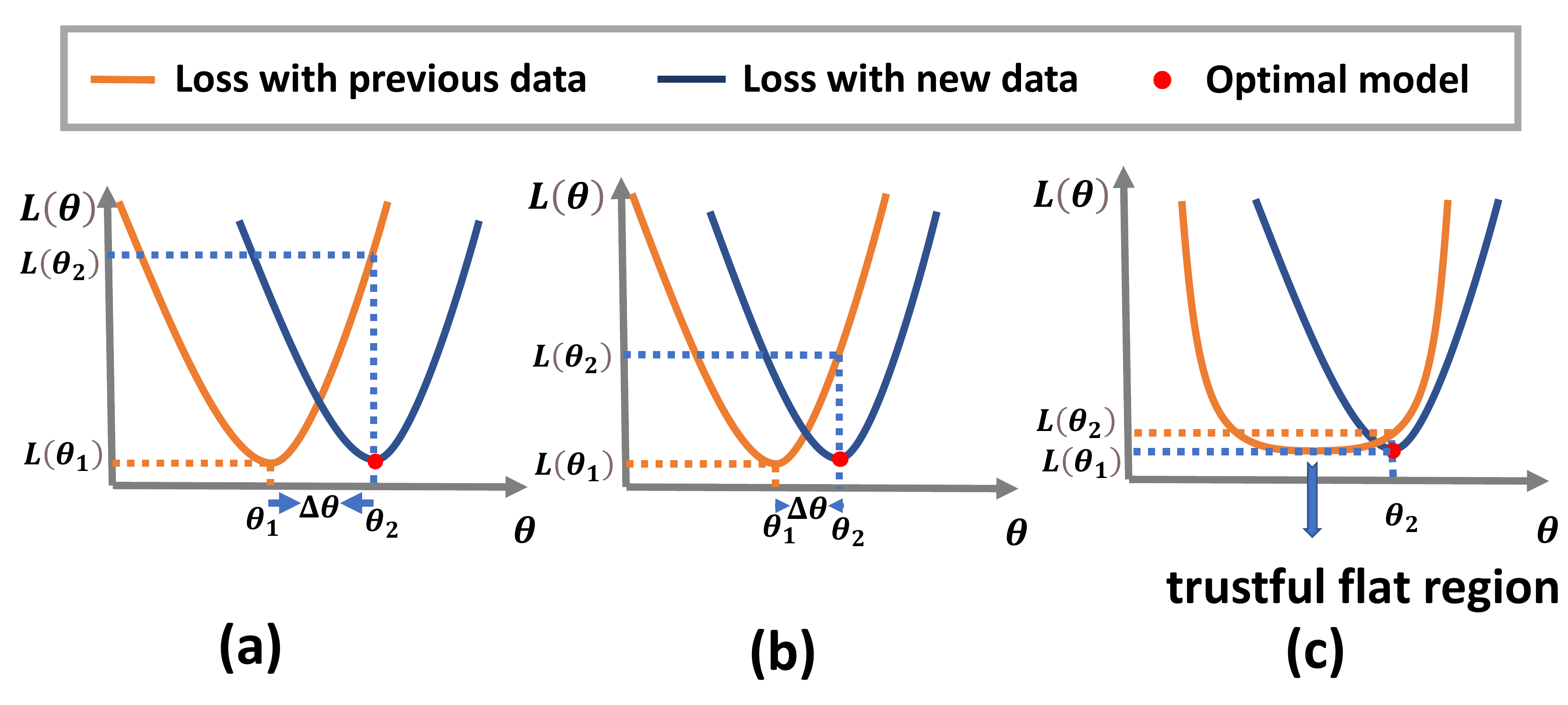}
 \vspace{-0.8cm}
  \caption{Motivation of the flat region.}
  \label{fig:flat_region}
  \vspace{-0.4cm}
\end{figure}
The key challenge of retraining the model is that, due to the privacy concerned short data retention policies in mobile and social applications and limited memory storage capacity of big stream data, a continuous learning model adapts to new data, new patterns, and new knowledge, while forgetting old data and knowledge at the same time.
Figure \ref{fig:flat_region} (a) shows after adapting to new data, the new model ($\theta_2$) shifts to the right side of the old model ($\theta_1$). There are two key observations of adapting to drifted data: 
1) the overlapped area between the new model and the old model becomes smaller and smaller.
2) the red inflection point of loss minimal in the new model results in a higher loss in the old model. 
Both observations indicate the new adapted model loses previous knowledge. 
How can we strive for a balance between anti-forgetting and adaptation?
Many studies develop various technical solutions \cite{liu2020learning} to impose a regularization: the parameters of the new model should not deviate too much from previous parameters, i.e.,
$\displaystyle min_{\theta_2} \|\theta_1 -\theta_2\|_2$.
Such a regularization term, however, can still cause a significant performance drop on previous data. Figure \ref{fig:flat_region} (b) shows the new model parameters are forced to be similar to the old model parameters, so that the overlapped area of the two models grows larger. 
However, the red-color loss minimal point of the new model still obtains a high loss in the old model, i.e., $\mathcal{L}(\theta_2) \gg \mathcal{L}(\theta_1)$. 

\noindent\textbf{Anti-forgetting Adaptation as A Minimax Game.} Inspired by \cite{schulman2015trust}, we leverage the concept of the flat region to strive a balance between anti-forgetting and adaptation in semi-supervised stream learning. 
Figure \ref{fig:flat_region} (c) shows why the flat region (denoted by $\theta^{\star} - a \leqslant \theta \leqslant \theta^{\star} + a$, where $a$ is the width of the flat region) concept works. 
Fundamentally, the flat region represents a set of optimal or near optimal candidate model instantiations in the model space on old data. 
Finding such a flat region can effectively increase the overlapped area of the new model and the old model and maintain a low loss on previous data, while at the same time allowing the new model to adapt and shift to new data. 
This flat region searching and optimization relaxation process to strive for a balance between anti-forgetting and adaptation can be described by finding a model parameter ($\theta$) that satisfies: 
\vspace{-0.2cm}
\begin{equation}
\begin{aligned}
\label{minmax}
 \min_{\theta} \quad &   \sum_{\mathbf{x}_i \in \textbf{D}_t} \mathcal{L}(\mathbf{x}_i;\theta) \\
    s.t. \quad &\theta^{\star} - a \leqslant \theta \leqslant \theta^{\star} + a.\\
\end{aligned}
\end{equation}
where the width $a$ of the flat region is manually specified based on empirical and domain experiences \cite{schulman2015trust,shi2021overcoming}. 

However, in a dynamic learning environment of SDSL, the best flat region width will dynamically vary when both old data and new data change at different times. 
Therefore, it is impractical to directly integrate the above formulation of the flat region. 
The key research question is: can we find the flat region while automatically identify the best width of the flat region?
We find that searching the flat region while identifying the best width can be reformulated into a computationally tangible minimax game. 
Assuming the parameters of a model can vary in a certain flat region defined by $\theta^{\star}-\xi \leqslant \theta  \leqslant \theta^{\star}  + \xi $.
We identify the upper bound of the model loss in the region of $\xi$ and find the worst case by maximizing the training loss of the network on the new data $\mathcal{\hat{D}}_t$, which is described by $max_{\xi} \sum_{\mathbf{x}_i \in \mathcal{D}_t } \mathcal{L}_{\mathcal{D}_{t}} (\mathbf{x}_i | \theta + \xi )$.
After the loss upper bound (worst case) over the region is measured, we minimize and lower the loss upper bound over the region to 
find the feasible parameters that can minimize the current loss.
Besides, as depicted in Figure \ref{fig:flat_region}.(c), $\xi$ lies in the neighborhoods around parameters train on previous data $\mathcal{D}_{t-1}$, the optimization of $\xi$ should follow $\xi \in \mathcal{M}$ where $\mathcal{M}$ represents the space span by previous parameters trained on $\mathcal{D}_{t-1}$. Formally, the objective function can be formulated as:
\begin{equation}
\begin{aligned}
\label{eq:minmax}
 \min_{\theta} \max_{\xi}\quad &  \sum_{\mathbf{x}_i \in \mathcal{D}_t } \mathcal{L}_{\mathcal{D}_{t}} (\mathbf{x}_i ; \theta + \xi )  \\
    s.t. \quad & \xi \in \mathcal{M}.\\
\end{aligned}
\end{equation}

\noindent\textbf{Solving the Optimization Problem.} Based on the gradient projection method \cite{rosen1960gradient}, the adversarial weight perturbation $\xi$ can be updated with the projection on space $M$ via the step size $\eta_{1}$,
\begin{equation}
    \xi \leftarrow  \xi + \eta_{1}
    \vectorproj[M](\nabla_{(\theta_t)} \mathcal{L}_{\mathcal{D}_t}(\theta_t +\xi)).
\label{eq:flat}
\end{equation}
Notably, $\theta$ at time $t-1$ preserves the previously acquired knowledge which is spanned by $M$. When coming to $\mathcal{D}_t$, We only update $\theta$ along the orthogonal direction of $M$, leading to the least change (or locally no change) to the learned $\theta$. Especially, the parameter $\theta$ can be adaptively updated with each $\mathcal{D}_t$ as:
\begin{equation}
    \theta \leftarrow  \theta - \eta_{2}(I-\vectorproj[\mathbf{M}])
     (\nabla_{(\theta_t)} L_{\mathcal{D}_t}(\theta_t +\xi)).
     \label{eq:theta}
\end{equation}
where $I$ is the identity matrix and $\eta_2$ is the step size. Concretely, We present the Algorithm overview of the model replay stage in Appendix A.2. Besides, we show a theoretical analysis on why the flat region can help mitigate forgetting when adapting on shifted streaming data with short lookback and why our method works in Appendix A.3.

\section{Experiments}
We conduct extensive experiments on various datasets to evaluate the performance of our method. 
Specifically, our experiments aim to answer the following questions:
\noindent\textbf{Q1:} Can our method outperform baselines on the semi-supervised drifted stream learning problem?
\noindent\textbf{Q2:} Can our  method generate robust pseudo-labels?
\noindent\textbf{Q3:} Can our method effectively alleviate the forgetting problem? 
\noindent\textbf{Q4:} Can the flat region theory be supported by empirical investigations?

\vspace{-0.2cm}
\subsection{Experimental Setup}
\subsubsection{Data Description}
We conducted experiments on eight datasets, including four widely-used synthetic benchmark datasets of stream learning research (i.e., UG\_2C\_2D, UG\_2C\_3D, UG\_2C\_5D and MG\_2C\_2D ) and four real-world stationary classification datasets (i.e., Optdigits, Spam, Satimage and Twonorm). 
Table \ref{Tab:statistics} shows the statistics of datasets. 
Specifically, we exploited the same setting in~\cite{souza2015data,guo2020record} to simulate the distribution shift manually by regrouping the instances for the four classification datasets, 
At each time, 1,000 instances arrived for UG\_2C\_2D and 2,000 instances arrived for UG\_2C\_3D, UG\_2C\_5D, MG\_2C\_2D, which were split into test and unlabeled datasets in a 30\% and 70\% ratio.  
For the Optdigits, Twonorm and Satimage datasets, 200 instances arrived each time and were split into 160 as unlabeled data and 40 as test data. 
For the Spam dataset, 400 instances arrived at every time step and were split into  280 as unlabeled data and 120 as test data. 
%Labeled data is available in the beginning and keep the same number with unlabeled data.

\begin{table}[h]
\vspace{-0.2cm}
\centering
\caption{Statistic Analysis of Datasets.}
\vspace{-0.2cm}
\begin{tabular}{cccc}
\hline
\textbf{Dataset} & Instances & Features &Classes\\
\hline
UG\_2C\_2D &  100,000&2 &2\\
UG\_2C\_3D & 200,000 &3 &2\\
UG\_2C\_5D & 200,000 &5&2 \\
MG\_2C\_2D&200,000&2&2\\
Optdigits&5620&64&10\\
Satimage &6435&36&7 \\
Spambase&9324&500&2\\
Twonorm & 7400 &2&2\\
\hline
\end{tabular}
\label{Tab:statistics}
\vspace{-0.3cm}
\end{table}

\begin{figure*}[h]
 \vspace{-0.2cm}
  \centering
  \subfigure[MG\_2C\_2D]{
  \includegraphics[width=0.23\textwidth,height = 0.13\textheight]{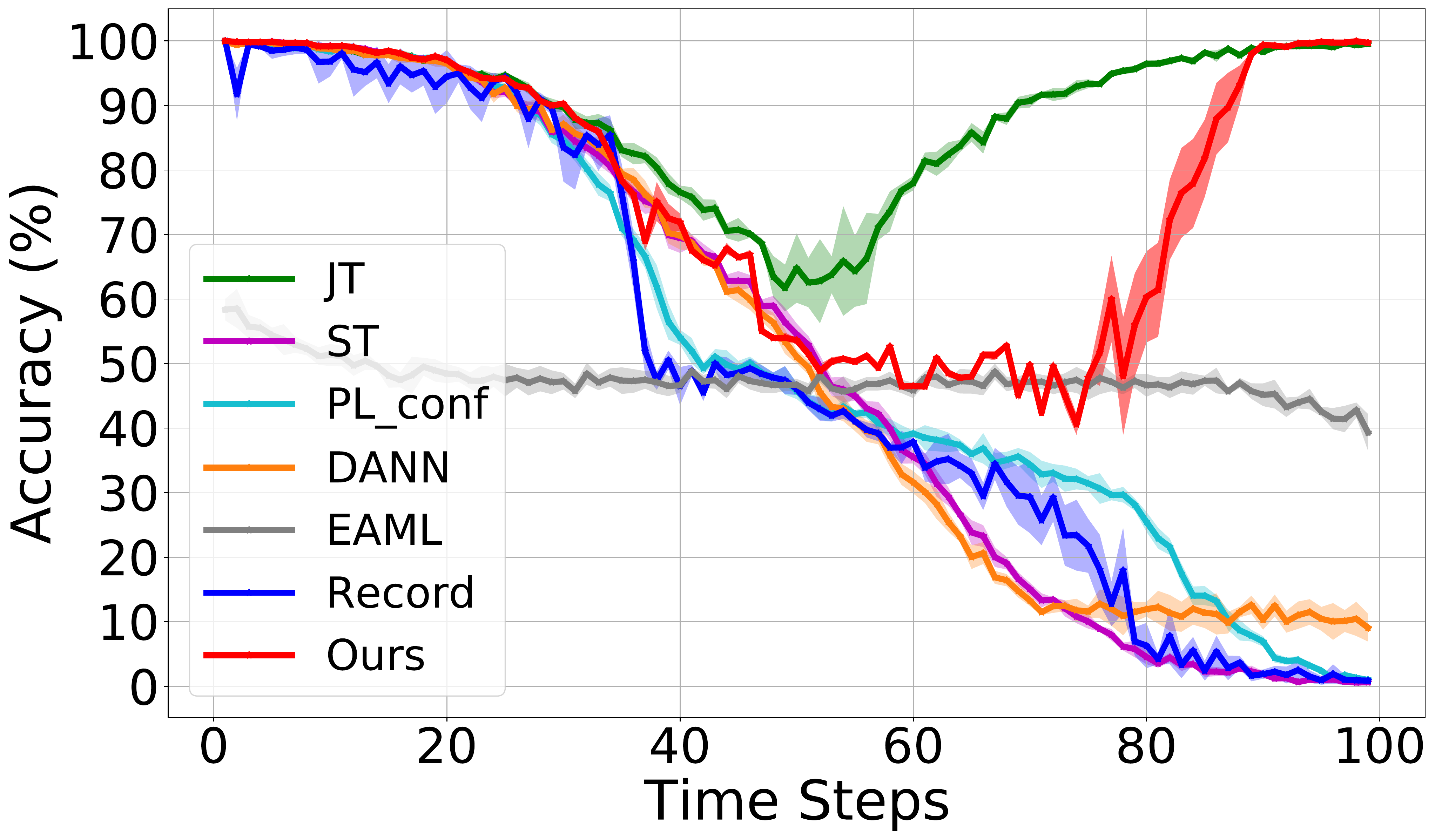}}
  % second
  \subfigure[UG\_2C\_2D]{
  \includegraphics[width=0.23\textwidth,height = 0.13\textheight]{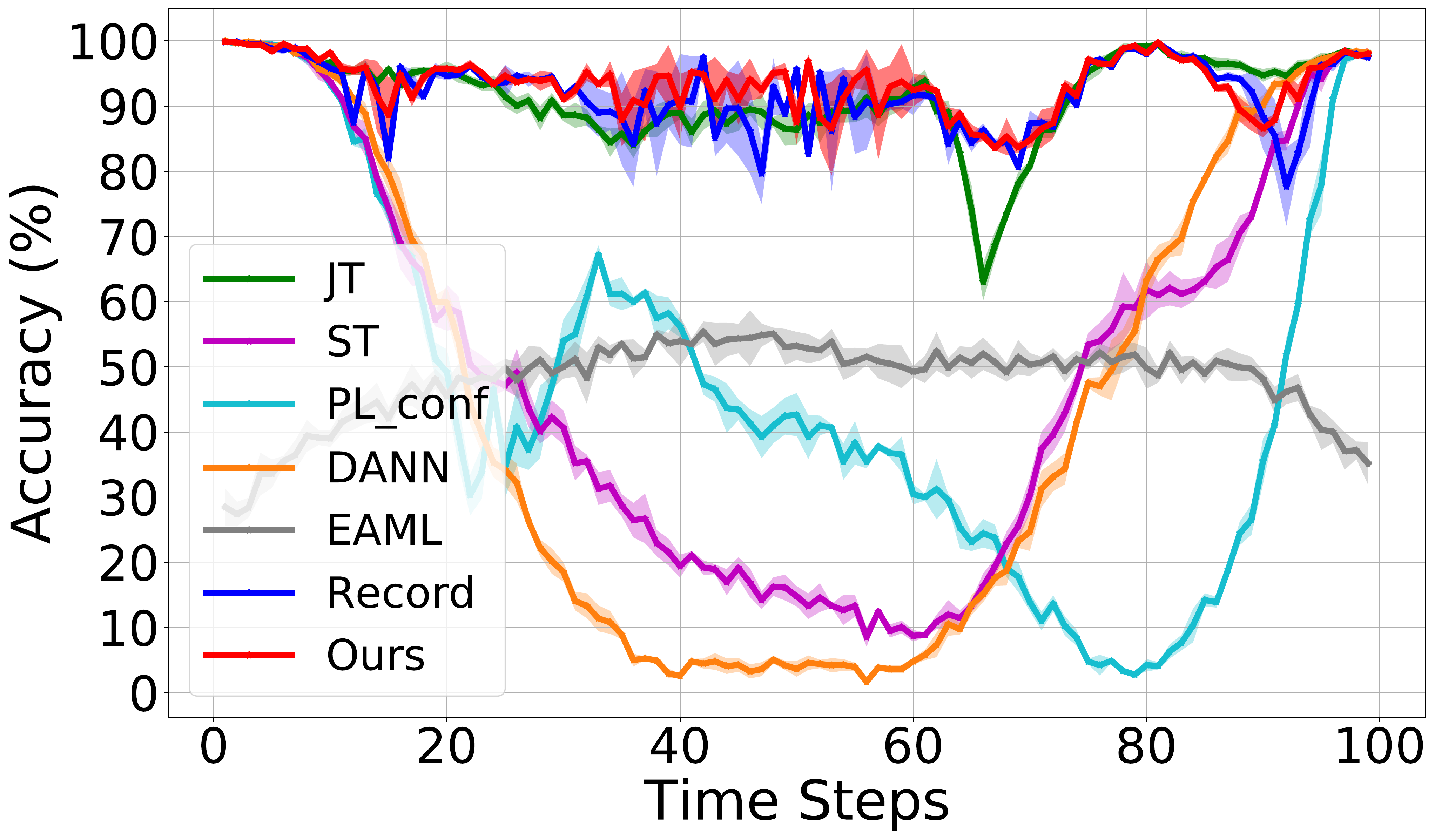}}
  % third
      \subfigure[UG\_2C\_3D]{
  \includegraphics[width=0.23\textwidth,height = 0.13\textheight]{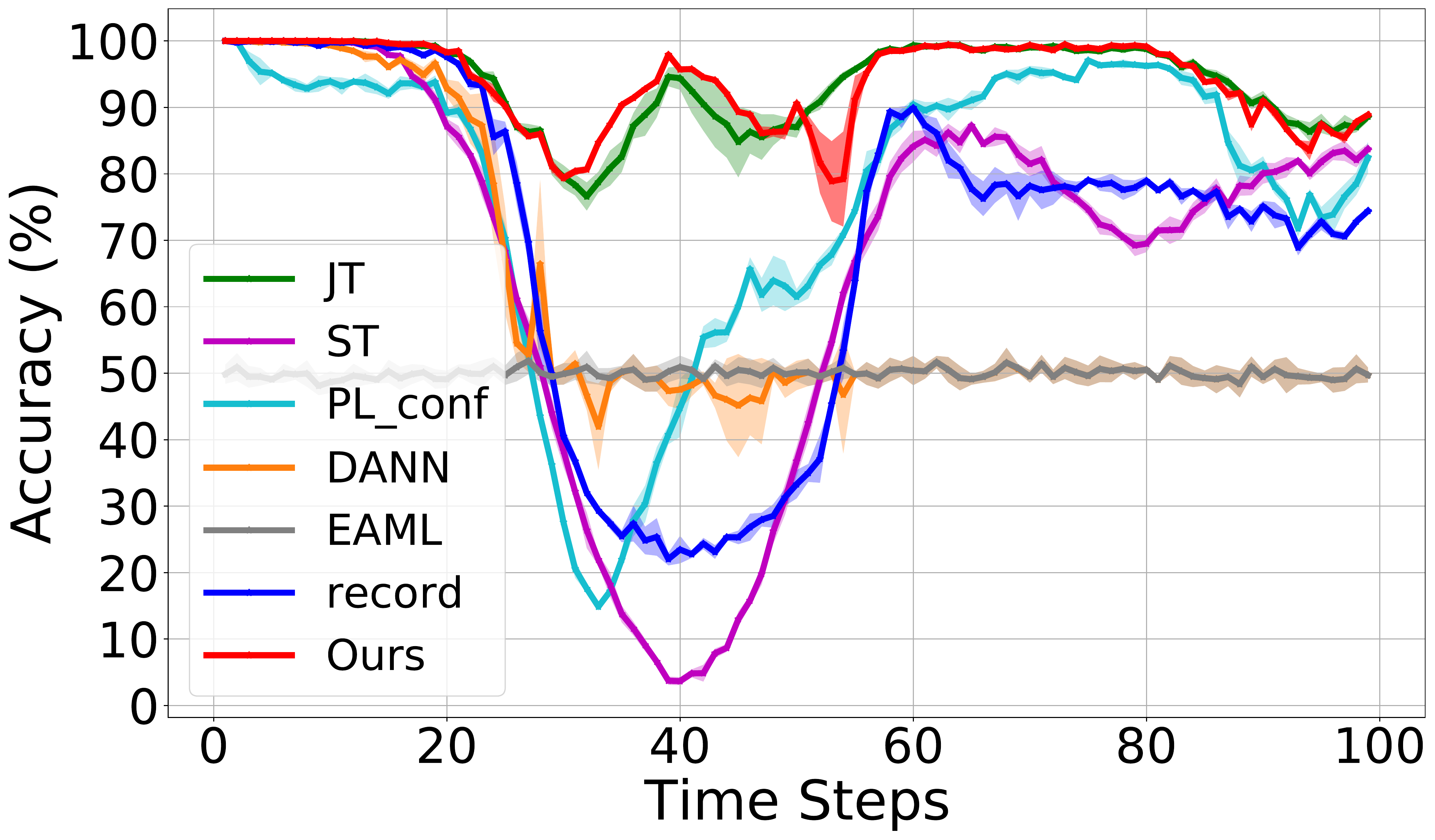}}
  %fourth
    \subfigure[UG\_2C\_5D]{
  \includegraphics[width=0.23\textwidth,height = 0.13\textheight]{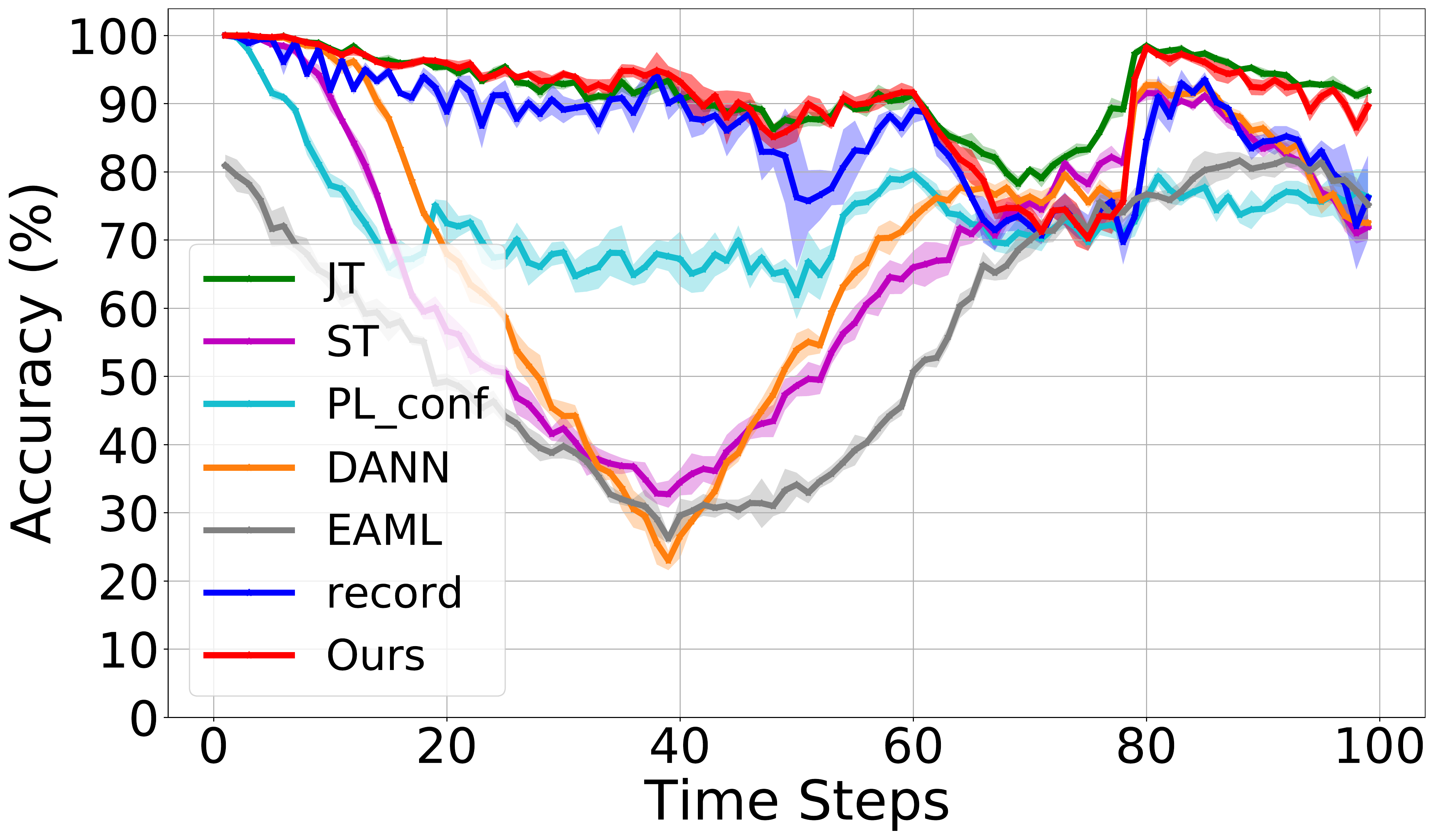}}
  %fifth
    \subfigure[Optdigits]{
  \includegraphics[width=0.23\textwidth,height = 0.13\textheight]{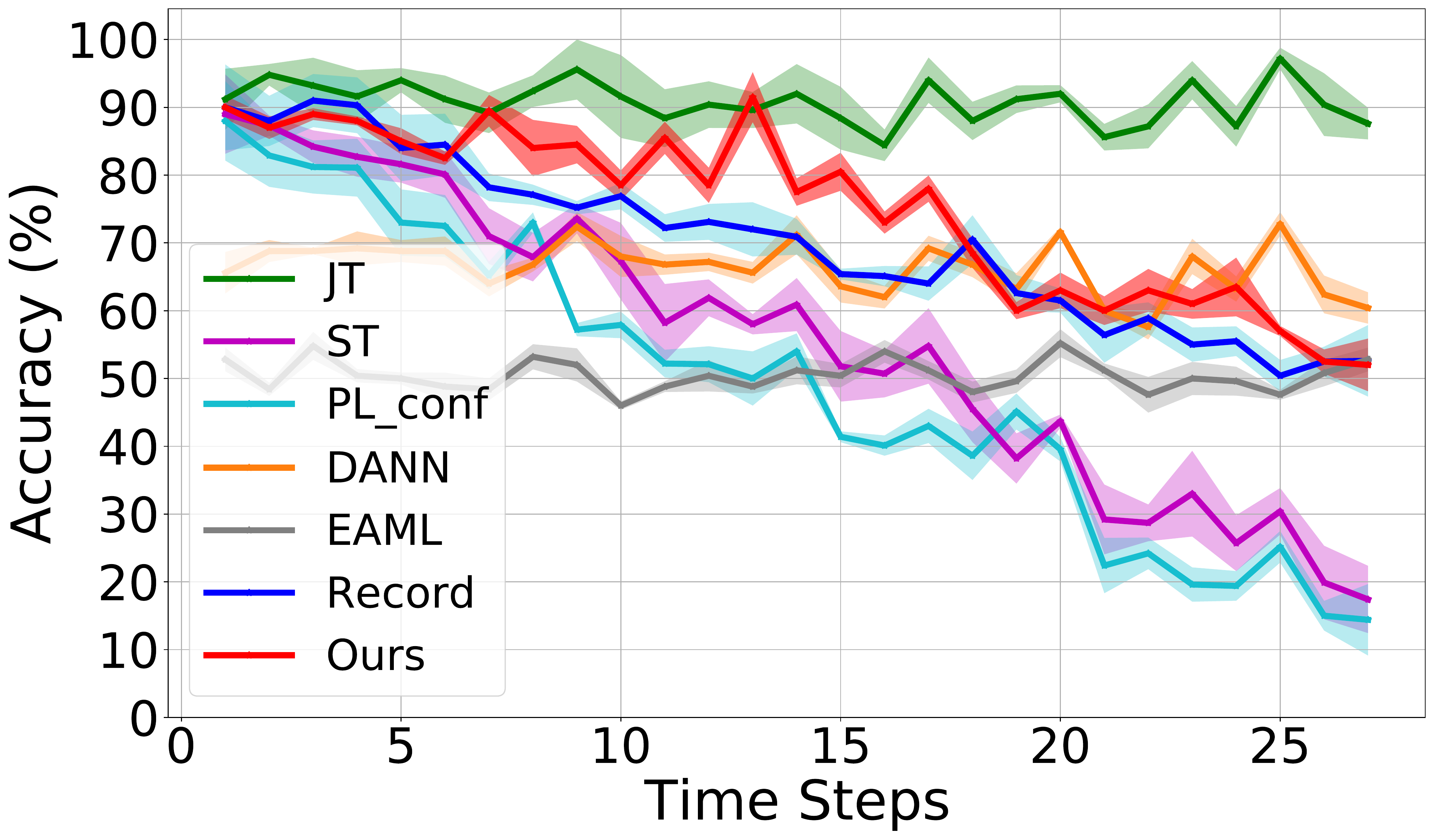}}
 %six 
    \subfigure[Satimage]{
  \includegraphics[width=0.23\textwidth,height = 0.13\textheight]{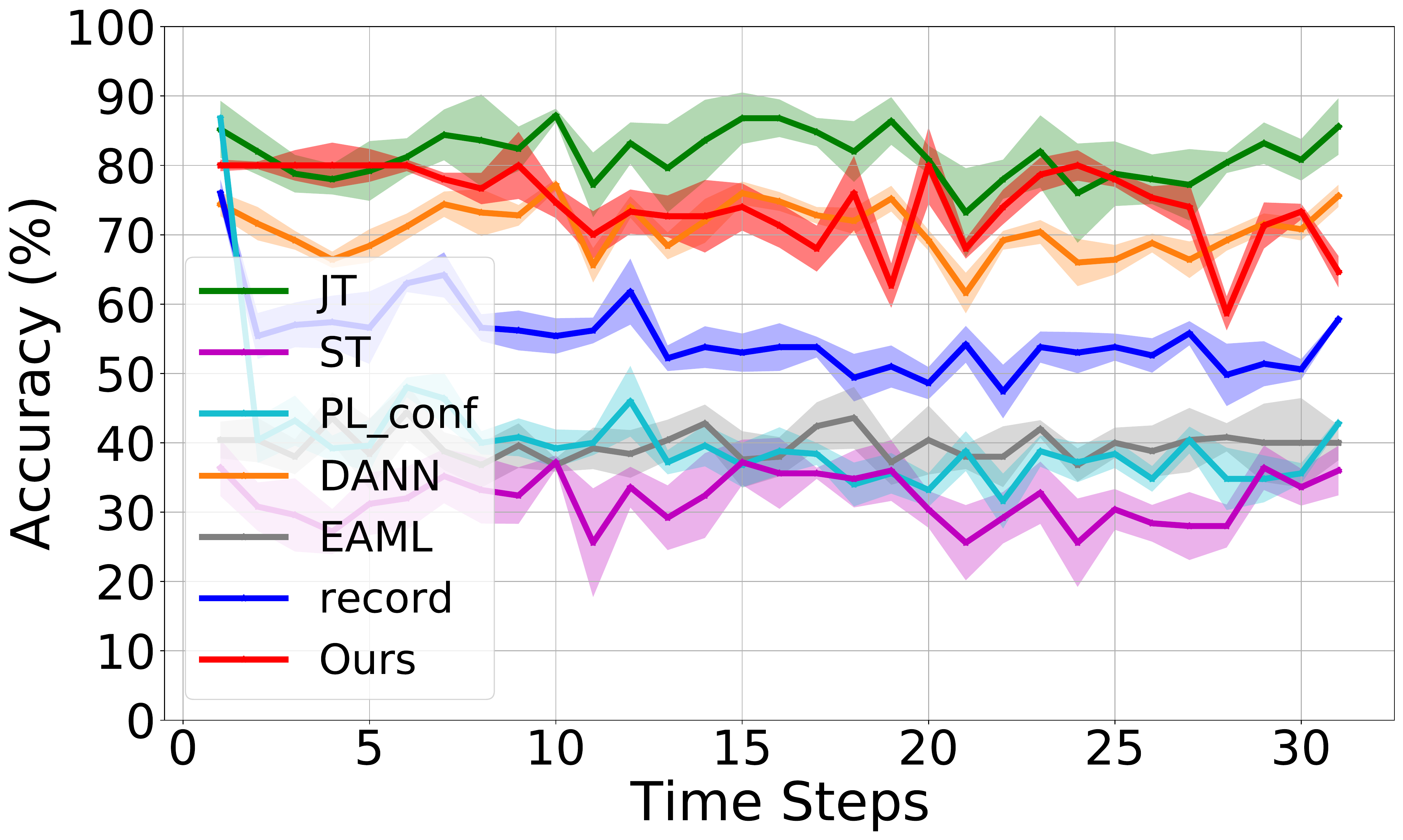}} 
  %seven
  \subfigure[Spambase]{
  \includegraphics[width=0.23\textwidth,height = 0.13\textheight]{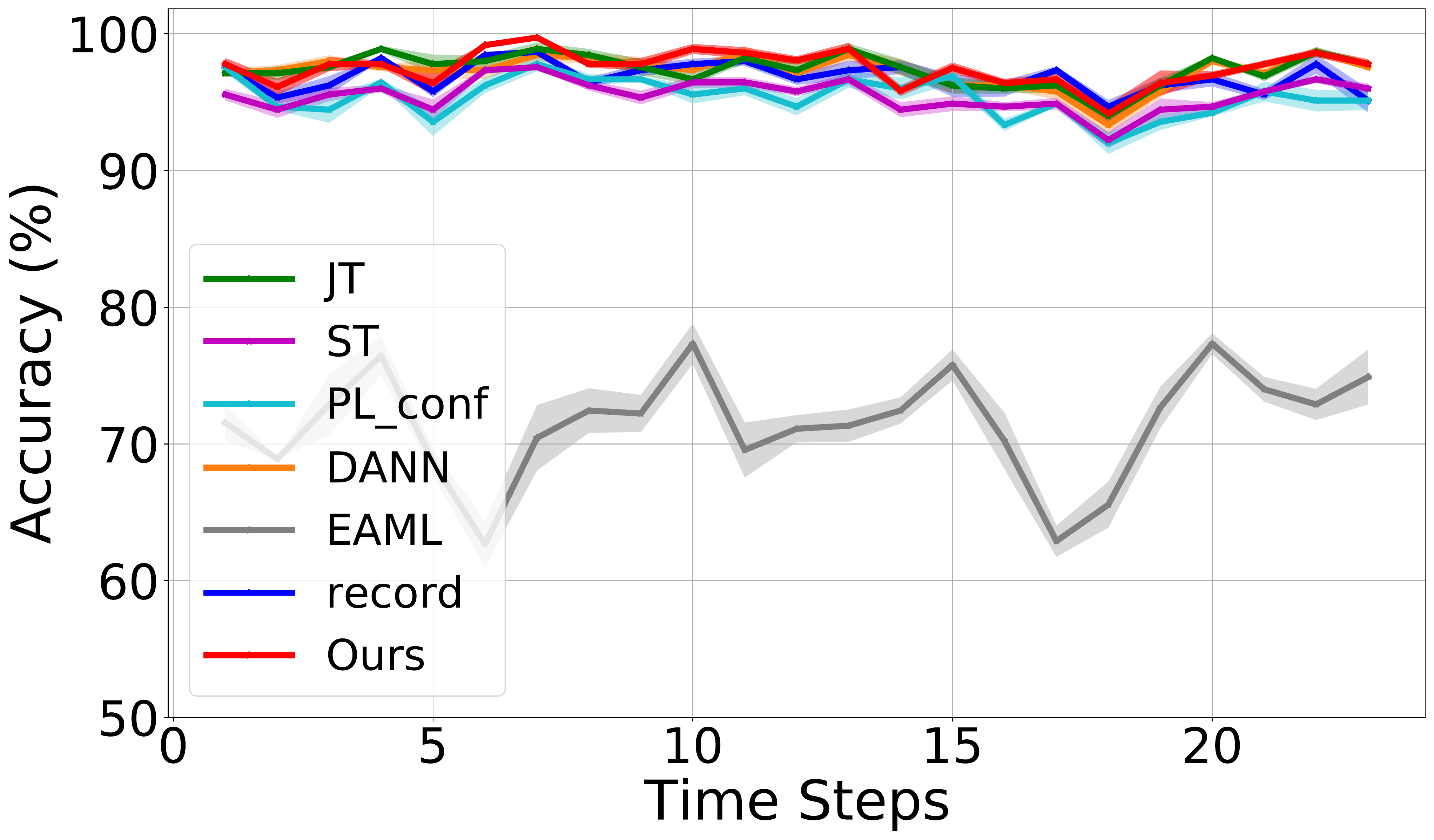}}
    \subfigure[Twonorm]{
  \includegraphics[width=0.23\textwidth,height = 0.13\textheight]{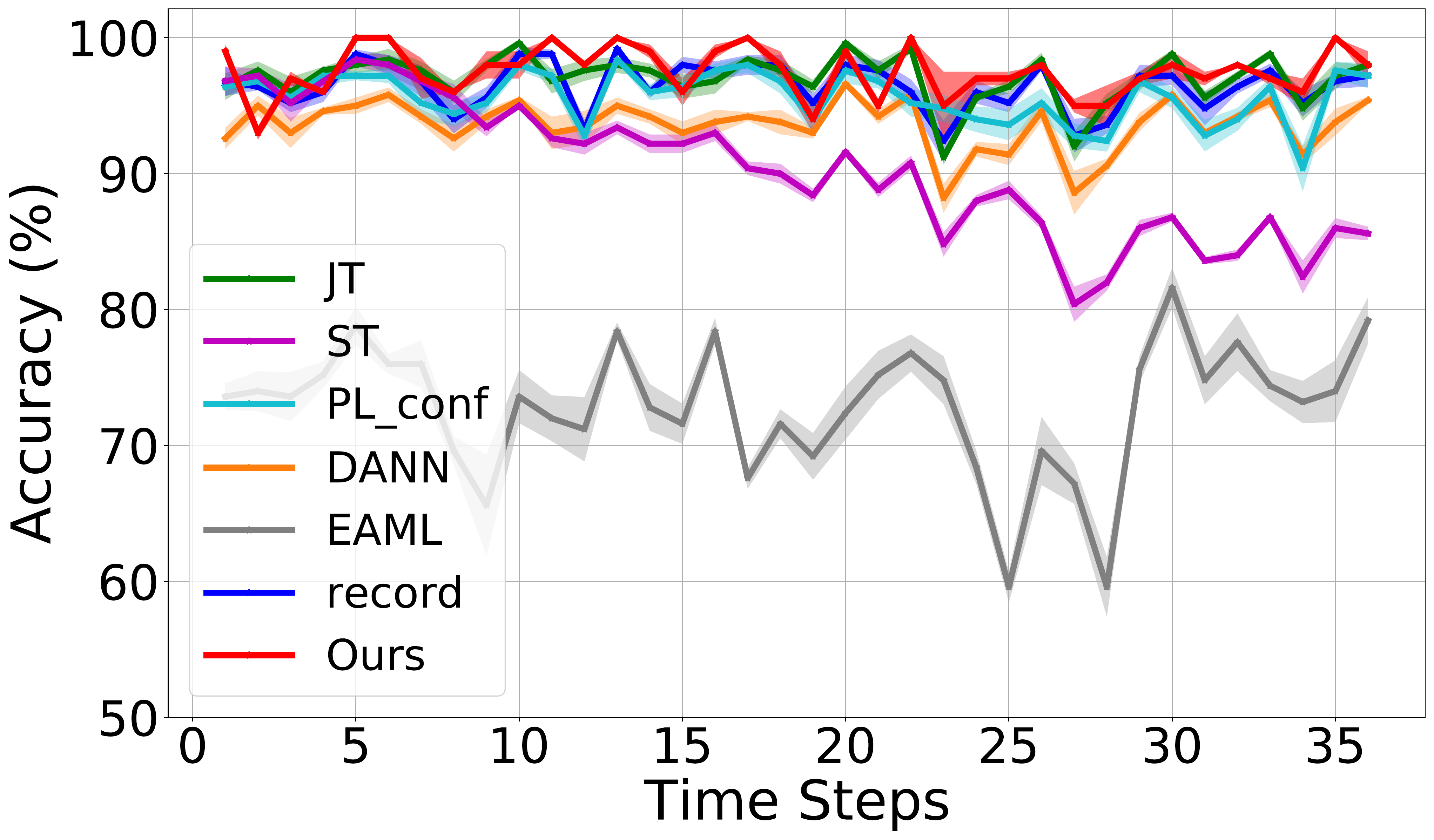}}
 \vspace{-0.5cm}
    \caption{Overall performance comparisons between our method and baseline methods.} \label{fig:case}
  \setlength{\abovecaptionskip}{0cm}
  \label{figure:1}
 \vspace{-0.2cm}
\end{figure*}

\subsubsection{Baseline Algorithms}
Since there are limited studies working on the SDSL setting, we compared our method with the semi-supervised learning methods~\cite{guo2020record}, domain adaptation methods via evolving shifted data~\cite{liu2020learning,tzeng2017adversarial} and other competitive baseline methods: 
\noindent(1) \textbf{Supervised Training (ST)}: simply trains the model on the gold standard labeled data once and test the model on stream data without adaptation. This setting can be viewed as the lower bound of the performance.
\noindent(2) \textbf{Joint Training (JT)}: assumes that the gold standard labels are available for all the streaming data at each time. The model is jointly trained on all the labeled data ever seen. JT is a strong baseline and can be viewed as an upper bound of the performance, since JT leverages all labeled data. 
\noindent(3) \textbf{Pseudo-labeling with high confidence (PL-Conf)} ~\cite{lee2013pseudo}:  stores examples with high softmax probability at each time.
\noindent(4) \textbf{Evolution Adaptive Meta-Learning (EAML)}~\cite{liu2020learning}: is a strong baseline that adapts to gradually shifted data without forgetting. EAML penalizes model parameters with a $L_2$ Regularizer to alleviate forgetting. Without access to incoming data labels, EAML minimizes feature discrepancy at different times as an alternative to cross-entropy loss.
\noindent(5) \textbf{Resource Constrained SSL under Distribution Shift (Record)} ~\cite{guo2020record}: exploits a generation-detection-restoring pipeline. Differently, Record needs to generate the pseudo- labeled set based on the previously trained model and restore influential samples ever seen with a memory buffer. 
\noindent(6) \textbf{Domain-adversarial training of neural networks (DANN)} ~\cite{tzeng2017adversarial}: is a representative domain adaptation method. We applied DANN to the evolving distribution shift setting by training the model with labeled data, and learning an invariant embedding on the evolving shifted data sequentially.

We reported the average results with standard deviations of 5 runs for all experiments.
Following the setting in~\cite{guo2020record}, we chose the mean teacher (MT) \cite{tarvainen2017mean} as the base model of SSL classifier in our framework. 
We used a two-layer multi-layer perceptron as a feature extractor. 
The memory reply buffer is set as 100 for our framework (the look back size) and baselines, i.e., only 100 unlabeled examples can be stored in memory. For PL\_Conf, we chose the 100 most confident samples to retrain the model.
The parameters of all the baseline models are defined in accordance with their respective publications. The step size $\eta_{1}$
and $\eta_{2}$ are set as 0.01.
\subsubsection{Evaluation Metric}
Following the setting in \cite{guo2020record, liu2020learning}, we evaluated the classification performance by averaging classification accuracy through each time as $Acc_t$:
\vspace{-0.3cm}
\begin{equation}
Acc_t = \frac{1}{T}\sum_{i=1}^{T}R_{i,i}.
\vspace{-0.25cm}
\end{equation}
And we evaluate the memorization ability by averaging classification accuracy on the final model as $Acc_T$:
\begin{equation}
Acc_T = \frac{1}{T}\sum_{i=1}^{T}R_{T,i}.
\vspace{-0.25cm}
\end{equation}
where $T$ is the total number of data sequences. $R_{i,j}$ is the test classification accuracy of the model at time $j$ after learning the last sample from $i$-th data. 
\vspace{-0.3cm}
%\subsection{Experimental Results}
\subsection{Q1: Overall Comparison}
%\subsubsection{Q1: Overall Comparison with Baselines}
To answer Q1, \textbf{we compared our method with baselines that leverage unlabeled data through different strategies.} 
Figure \ref{figure:1} shows the mean classification accuracy and standard deviation on the test data for five runs. The Y-axis represents accuracy ($Acc_t$ here), while the shaded regions show standard error computed using various random seeds. Figure~\ref{figure:1} shows that our method achieves significant improvements over the baselines and is even comparable with the JT method (upper bound of the setting). 
%This is because our framework leverages labeled data $\mathbf{D}$ and a short lookback pseudo label set $\mathbf{D}_{t-1}$. The lookback data and adaptive Nearest Neighbor classifier both help to generate high-quality pseudo labels.
The observation validates the effectiveness of leveraging multi-level knowledge (supervised, unsupervised, and structure) for robust labeling and minimax-based flat region solver for anti-forgetting adaptation. 

Besides, one interesting observation arises from the results that unlabeled data matters for streaming adaption in the semi-supervised setting. 
Specifically, both DANN and EAML perform poorly in streaming shifted data.  Our method and Record that utilize pseudo labels show significantly better performance. This implies pseudo labels can introduce auxiliary information about shifted data and improve generalization ability for streaming data. 
Moreover, while Record benefits from the influential shifted data detection mechanism, the pseudo labels are generated by the classifier trained on previous data. 
The performance gap between our method and Record validates the necessity to mitigate classifier bias and verified our motivation to leverage the structure knowledge of invariant label class semantics.
% Note that the outstanding performance of our method only depends on limited labeled data, i.e. golden data and a short look back and current unlabeled data, indicates that our model can quickly learn a usable model rather than waiting for more data to be collected. 
\vspace{-0.3cm}
\subsection{Q2: Study of Robust Pseudo Labeling}
\iffalse
\subsubsection{Effectiveness in Robust Pseudo Labeling }
\begin{figure}[h]
  \vspace{-0.5cm}
  \centering
  \subfigure[UG\_2C\_5D]{
  \includegraphics[width=0.23\textwidth,height = 0.13\textheight]{samples/figure/UG_2C_5D_diff_prob_30.pdf}}
  % second
  \subfigure[Satimage]{
  \includegraphics[width=0.23\textwidth,height = 0.13\textheight]{samples/figure/satimage_arff_diff_prob_100.pdf}}
%   \caption{classification results w/o robust labeling component on UG\_2C\_5D and Satimage dataset through each round.}
\caption{Validation of robust pseudo labeling.}
    % \vspace{-0.5cm}
    \label{Fig:robust_labeling}
 \vspace{-0.5cm}
\end{figure}
\fi

\subsubsection{Effectiveness of Robust Pseudo Labeling }
% To understand the impact of robust pseudo labeling strategy, we test our method with two settings for which pseudo labels are generated. Ours-PL, ours-Record that replaces the generation strategy with PL\_conf and Record method, respectively. 
To validate the effectiveness of the proposed robust pseudo labeling method, we compare our method with two variants. 
Since our proposed solution is a generation-replay pipeline, we replace the generation part (pseudo label generation) with two widely used methods to construct the two variants: (1) the variant that takes PL\_conf as the generation part, denoted by ``Ours\_PL''; 
(2) the variant that takes the Record's generation method as the generation part, denoted by ``Ours\_Record''. 
The difference is the variant ``Ours\_PL'' selects the most confident samples based on the softmax-based predictive confidence, while the variant ``Ours\_Record'' selects the influential samples for the pseudo-labels generation.

% To understand the impact of robust pseudo labeling strategy, we test our method with two settings for which pseudo labels are generated. Ours-PL, Ours-Record that replaces the generation strategy with PL\_conf and Record method, respectively.
% %Figure \ref{Fig:robust_labeling} summarizes the results on UG\_2C\_2D and Satimage dataset.
% \textbf{Both record and PL\_conf generates pseudo labels based on previously trained model. The difference lies in the selection strategies.
% Ours\_record replaces the generation strategy with record method, which they track the distribution changes and locate the distribution shifted samples. Then the most influential sample for the distribution change is selected based on a influence- based approach.
% PL\_conf selects the most confidence samples based on softmax value.}

\begin{figure}[h]
  \vspace{-0.5cm}
  \centering
  \subfigure[Classification performance through timelines on the UG\_2C\_5D dataset. ]{
  \includegraphics[width=0.23\textwidth,height = 0.13\textheight]{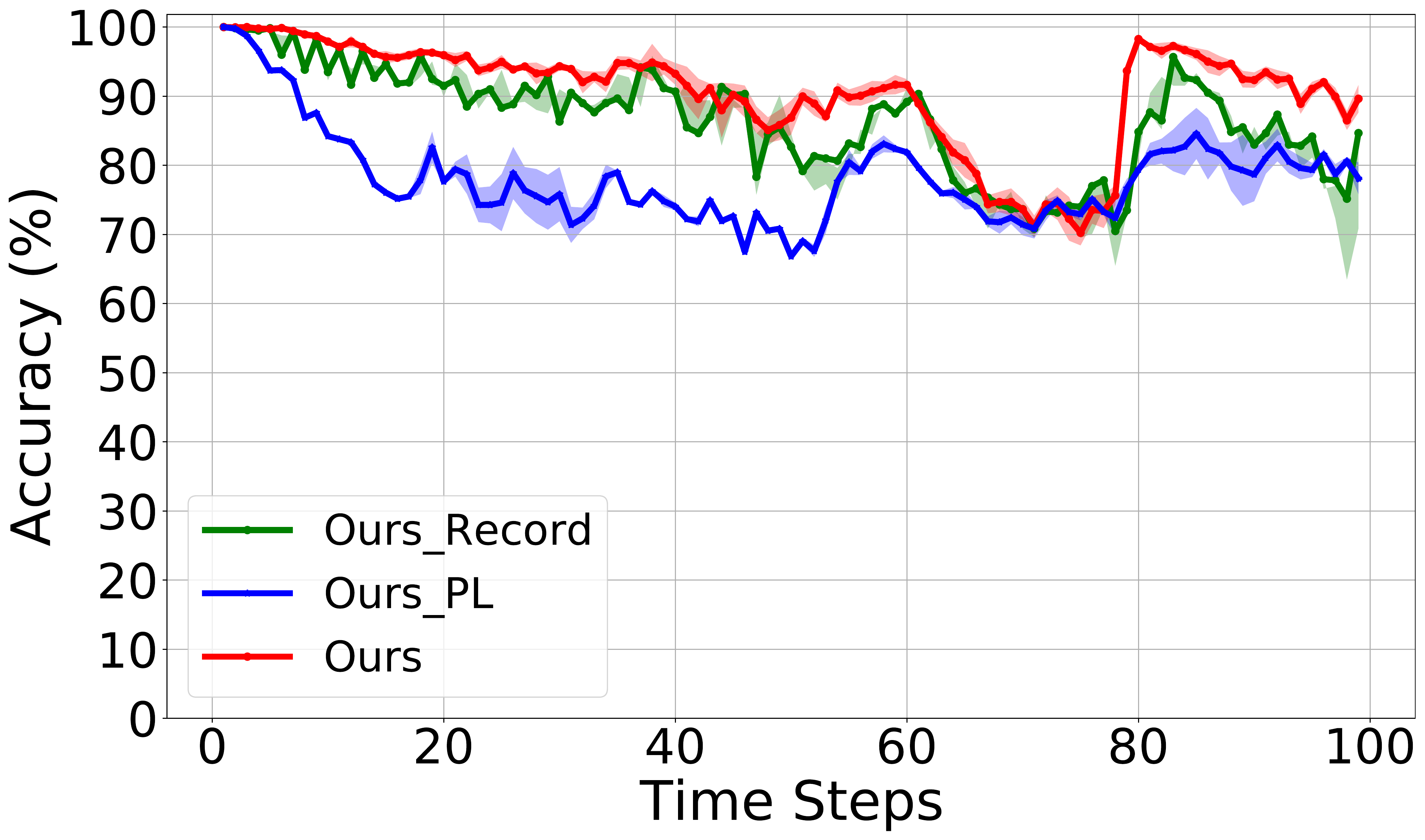}}
  % second
  \subfigure[Classification performance through timelines on the Satimage dataset.]{
  \includegraphics[width=0.23\textwidth,height = 0.13\textheight]{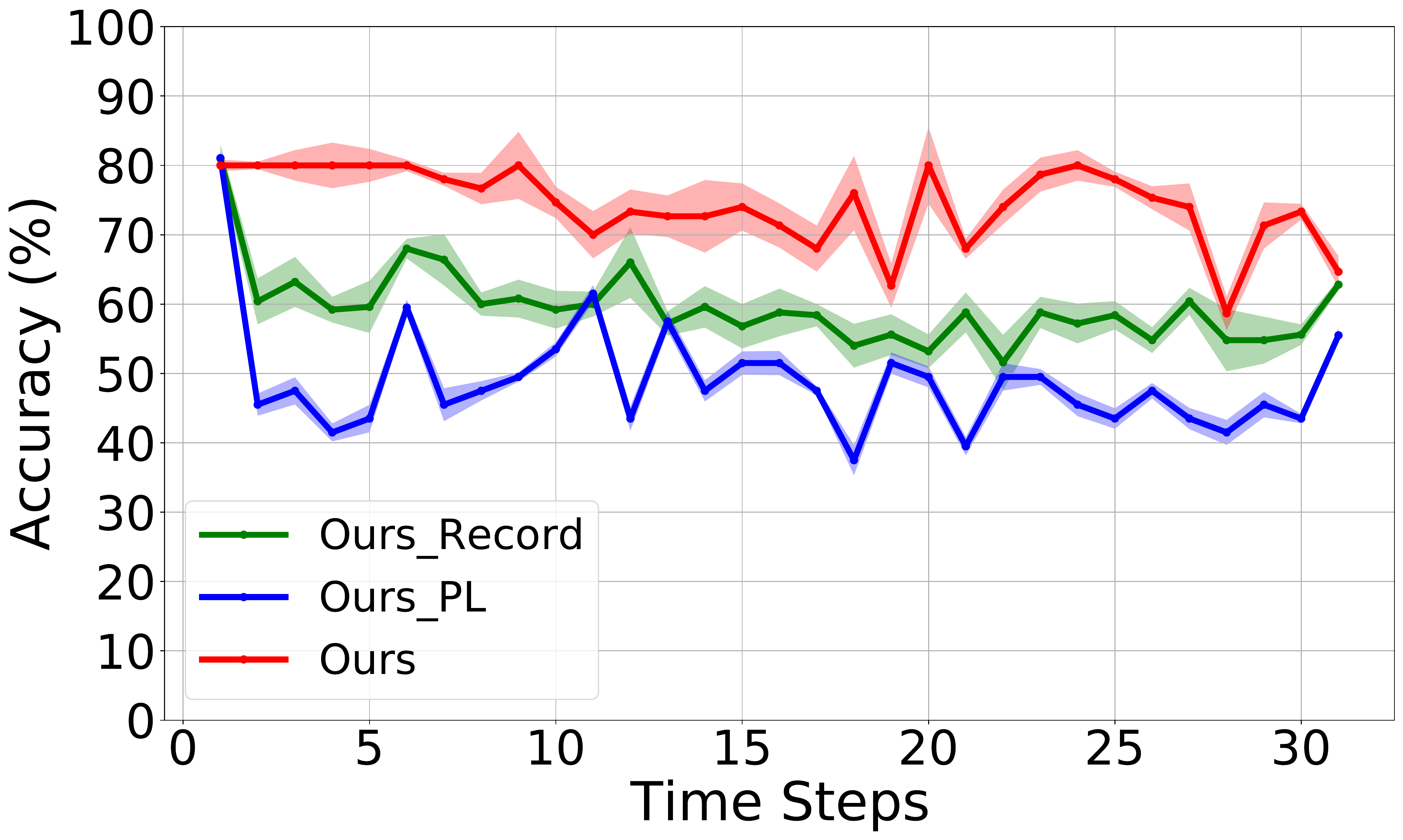}}
%   \caption{classification results w/o robust labeling component on UG\_2C\_5D and Satimage dataset through each round.}
\vspace{-0.4cm}
\caption{Validation of robust pseudo labeling.}
\vspace{-0.4cm}
\label{Fig:431_robust_pseudo_labeling}
\end{figure}

Figure~\ref{Fig:431_robust_pseudo_labeling} shows our method uses the robust pseudo labeling method and outperforms the two variants. Specifically, the variant ``Ours\_Record'' performs better than the variant ``Ours\_PL'' since the Record method selects the most influential samples for the distribution change. 
This observation verified our motivation that mining unlabeled shifted data could boost adaptation performance. 
However, the selected pseudo labels still lie in the overlap region with previous data. And it is insufficient to incorporate the non-overlapped information.
In contrast, our method achieves consistently promising performance. 
The boost in performance verifies our motivation that class centroid-based clustering can exploit the global pattern structure and assign accurate labels for the new unlabeled drifted data that are non-overlapped with previous knowledge.
  \vspace{-0.4cm}
\vspace{-0.4cm}

\subsubsection{Ablation study of Robust Pseudo labeling} 
Our proposed robust pseudo labeling method generates promising pseudo labels for unlabeled data by preserving the invariant label semantics.
To validate the contribution of the invariant label semantics, we conduct the ablation study, in which we compare our method with the variant that omits the Invariant Label Semantics (ILS) constraint in Equation~\ref{Eq:semantics}. 
We denote the variant as ``Ours w/o ILS''.

\begin{figure}[h]
  \vspace{-0.3cm}
  \centering
  \subfigure[Classification performance ($Acc_t$) through timelines on Satimage data.]{
  \includegraphics[width=0.22\textwidth,height = 0.13\textheight]{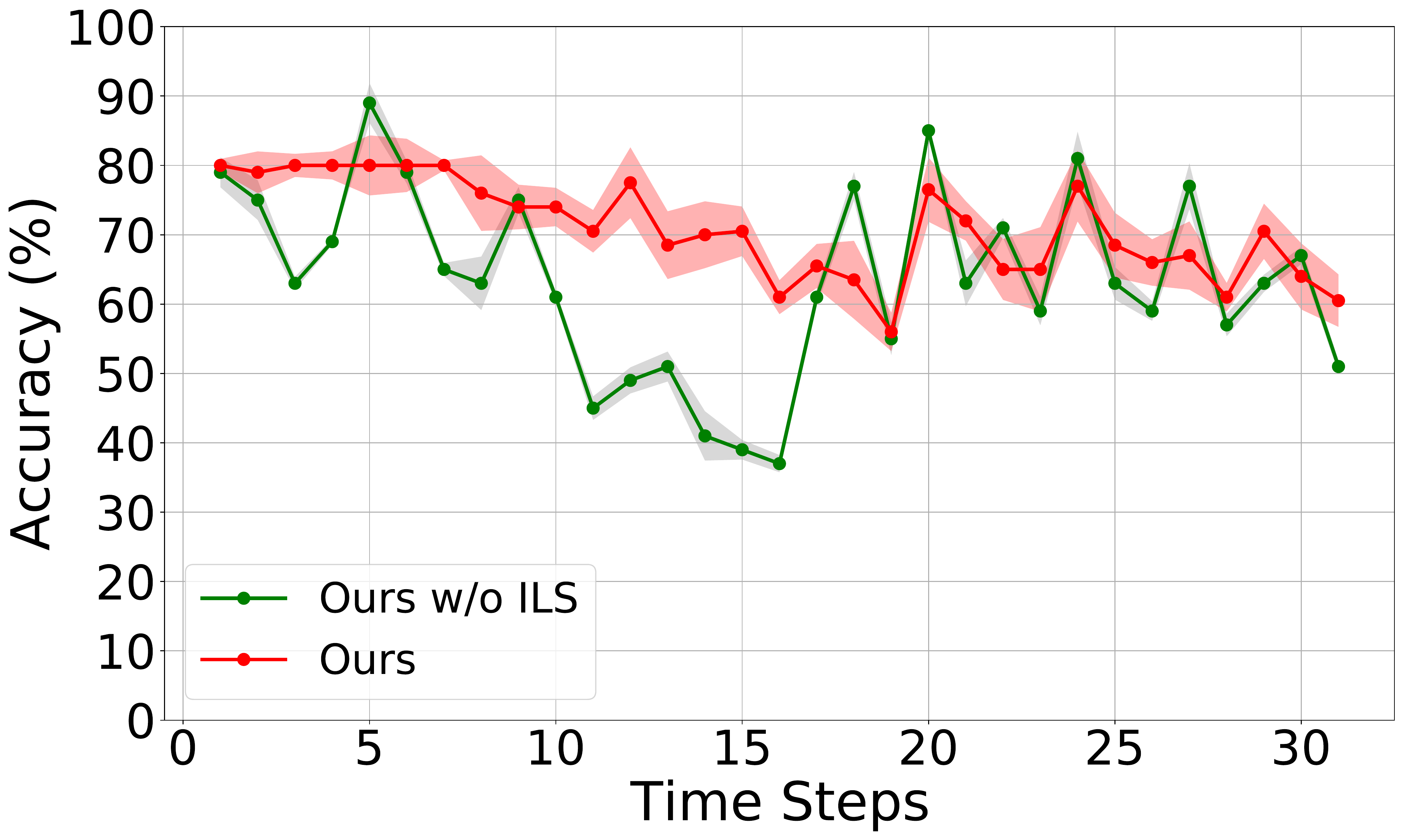}}
  \subfigure[Memorization ability ($Acc_T$) through timelines on Satimage data.]{
  \includegraphics[width=0.22\textwidth,height = 0.13\textheight]{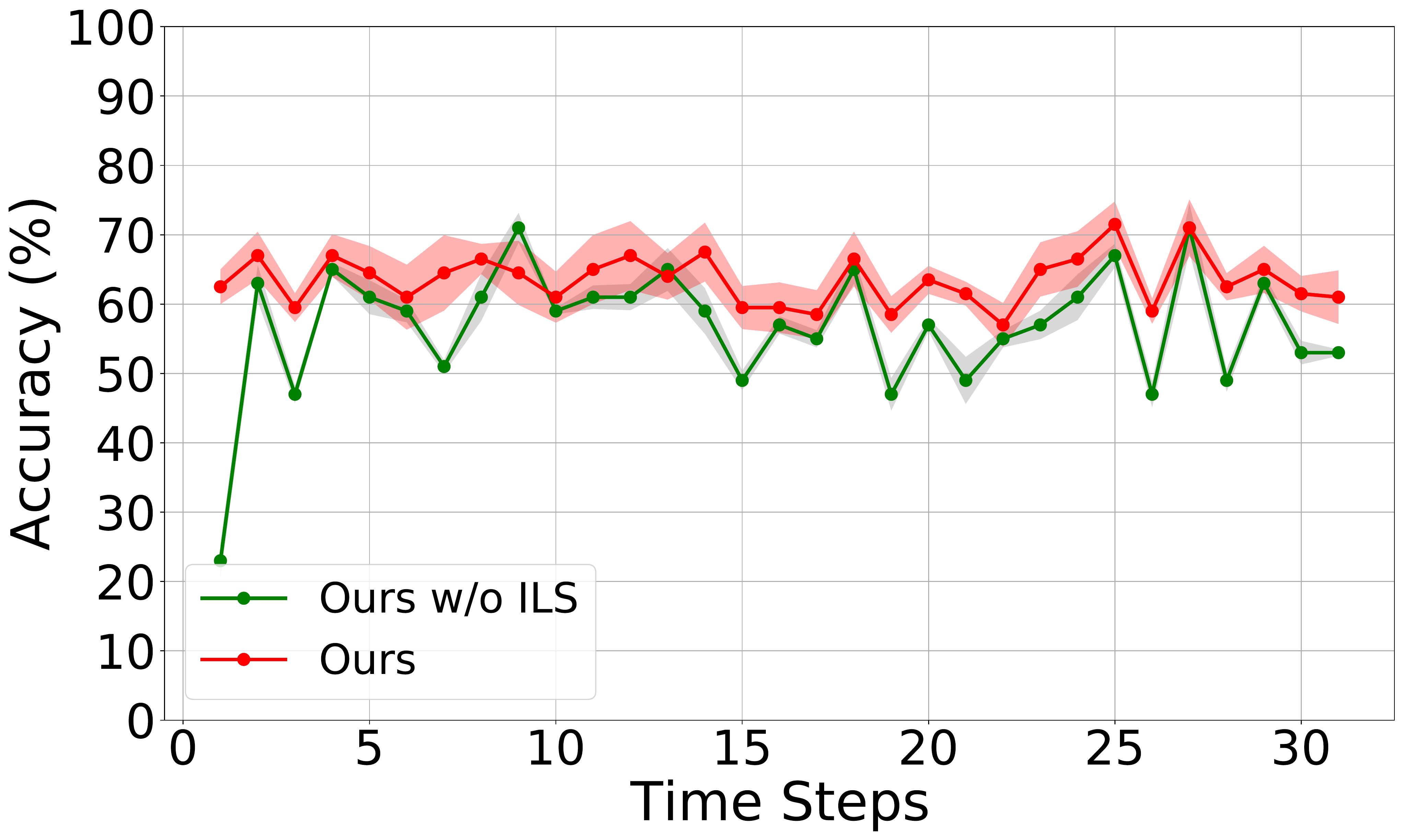}}
\vspace{-0.3cm}
\caption{Ablation study in Pseudo labeling.}
\vspace{-0.5cm}
\label{Fig:robust_labeling_ablation}
\end{figure}

Figure~\ref{Fig:robust_labeling_ablation} suggests that the ILS constraint contributes a stable classifier learning and can generate high-quality pseudo-labeled pairs. so, the model is easy to adapt well and mitigate forgetting to some extent due to the denoise ability. 
The observation verifies our motivation that the utilization of unsupervised knowledge of new data could provide extra information than previous labeled data and benefits the pseudo label generation of shifted data.
 
\subsection{Q3: Study of Alleviating Forgetting}
\begin{table*}[t!]
\centering
% \caption{End task performance on previously seen data over all baseline methods. High value indicates the model keeps high competence over previous knowledge. Joint Training (JT) shows upper bound performance.}
\caption{Comparison of effectiveness on alleviating forgetting. Noted that Joint Training (JT) shows the ideal performance of the SDSL setting. The closer to JT, the better the performance.}
\begin{tabular}{cllllllll}
\hline
\textbf{Method} & MG\_2C\_2D & UG\_2C\_2D&UG\_2C\_3D& UG\_2C\_5D&Optdigits
&Satimage&Spam&Twonorm\\
\hline
ST 
& 0.479$\pm$0.020
&0.367$\pm$0.020 
& 0.438 $\pm$0.019
&0.570$\pm$0.015
& 0.439$\pm$0.032
&0.318$\pm$0.019
&0.965$\pm$0.018
&0.881$\pm$0.009\\

JT & 0.593$\pm$0.012 
&0.907$\pm$0.0252 
&0.837$\pm$0.0103
&0.919$\pm$0.007
&0.921$\pm$0.019
&0.816 $\pm$0.018
&0.971$\pm$0.016 
&0.966$\pm$0.012\\

PL\_conf 
&0.513$\pm$0.020
&0.531$\pm$0.021
&0.567$\pm$0.219
&0.614$\pm$0.019
&0.406$\pm$0.025
&0.385$\pm$0.026
&0.966$\pm$0.021
& 0.956$\pm$0.011\\

DANN 
&0.532$\pm$0.039
&0.349$\pm$0.011
&0.578$\pm$0.018
&0.701$\pm$0.014
&0.415$\pm$0.031
&0.607$\pm$0.019
&0.969$\pm$0.013
&0.948$\pm$0.018\\

EAML 
&0.508$\pm$0.014
&0.499$\pm$0.016
&0.499$\pm$0.012
&0.569$\pm$0.016
&0.527$\pm$0.018
&0.397$\pm$0.015
&0.725$\pm$0.014
&0.729$\pm$0.021\\

Record 
&0.499$\pm$0.021
&0.883$\pm$0.010
&0.599$\pm$0.035
&0.725$\pm$0.020
&0.613$\pm$0.021
&0.536$\pm$0.026 
&0.965$\pm$0.018
&0.961$\pm$0.010 \\

Ours
&0.549$\pm$0.021
&0.894$\pm$0.017
&0.717$\pm$0.015
&0.768$\pm$0.057
&0.657$\pm$0.024
&0.635$\pm$0.034
&0.973$\pm$0.012 
&0.964$\pm$0.015 \\
\hline
\end{tabular}
\label{Tab:forget_competence}
\end{table*}

\begin{figure}[t]
  \centering
  \subfigure[Memorization ability through timelines on the UG\_2C\_5D dataset.]{
  \includegraphics[width=0.23\textwidth,height = 0.13\textheight]{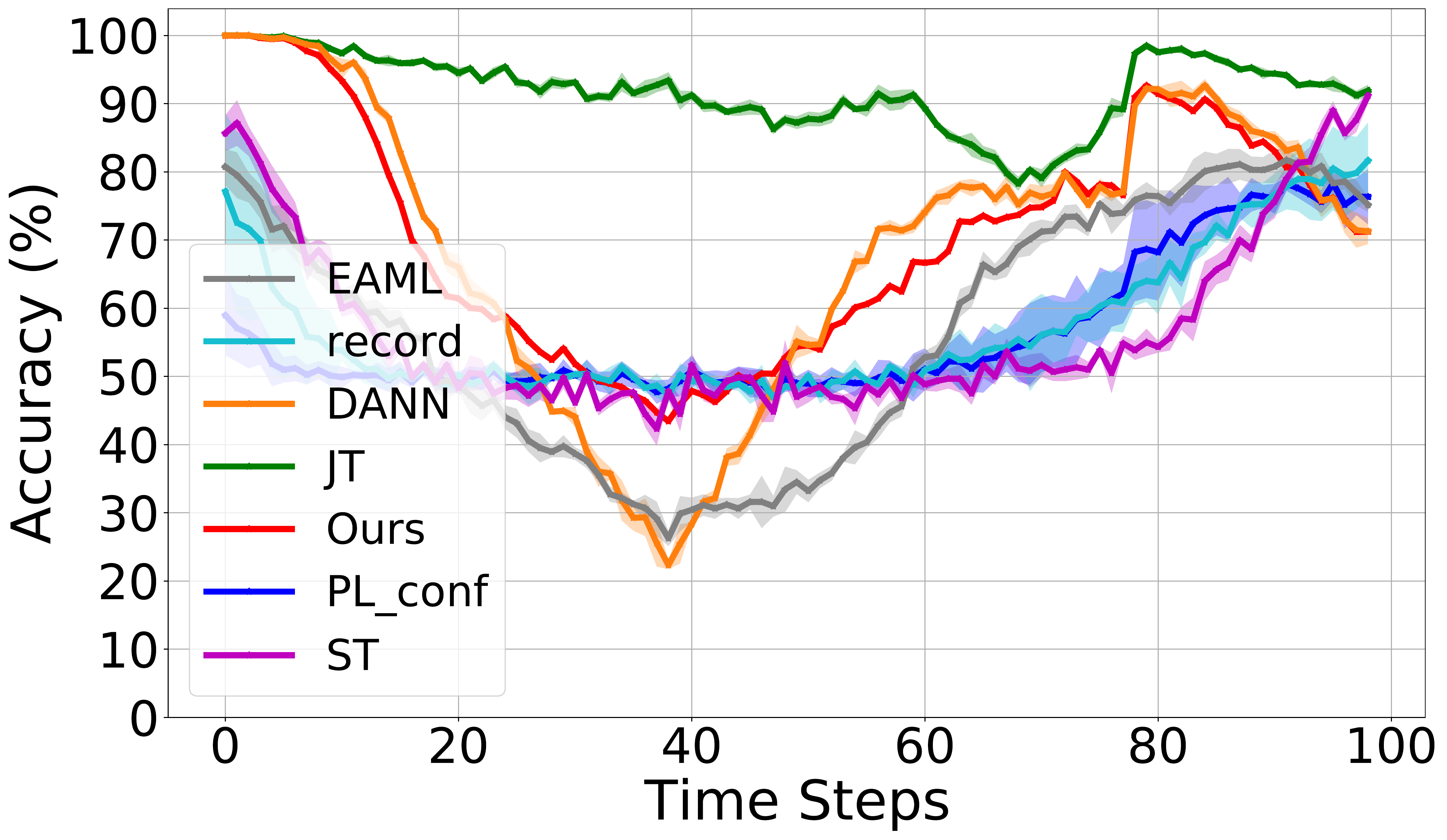}}
  % second
  \subfigure[Memorization ability through timelines on the Satimage dataset.]{
  \includegraphics[width=0.23\textwidth,height = 0.13\textheight]{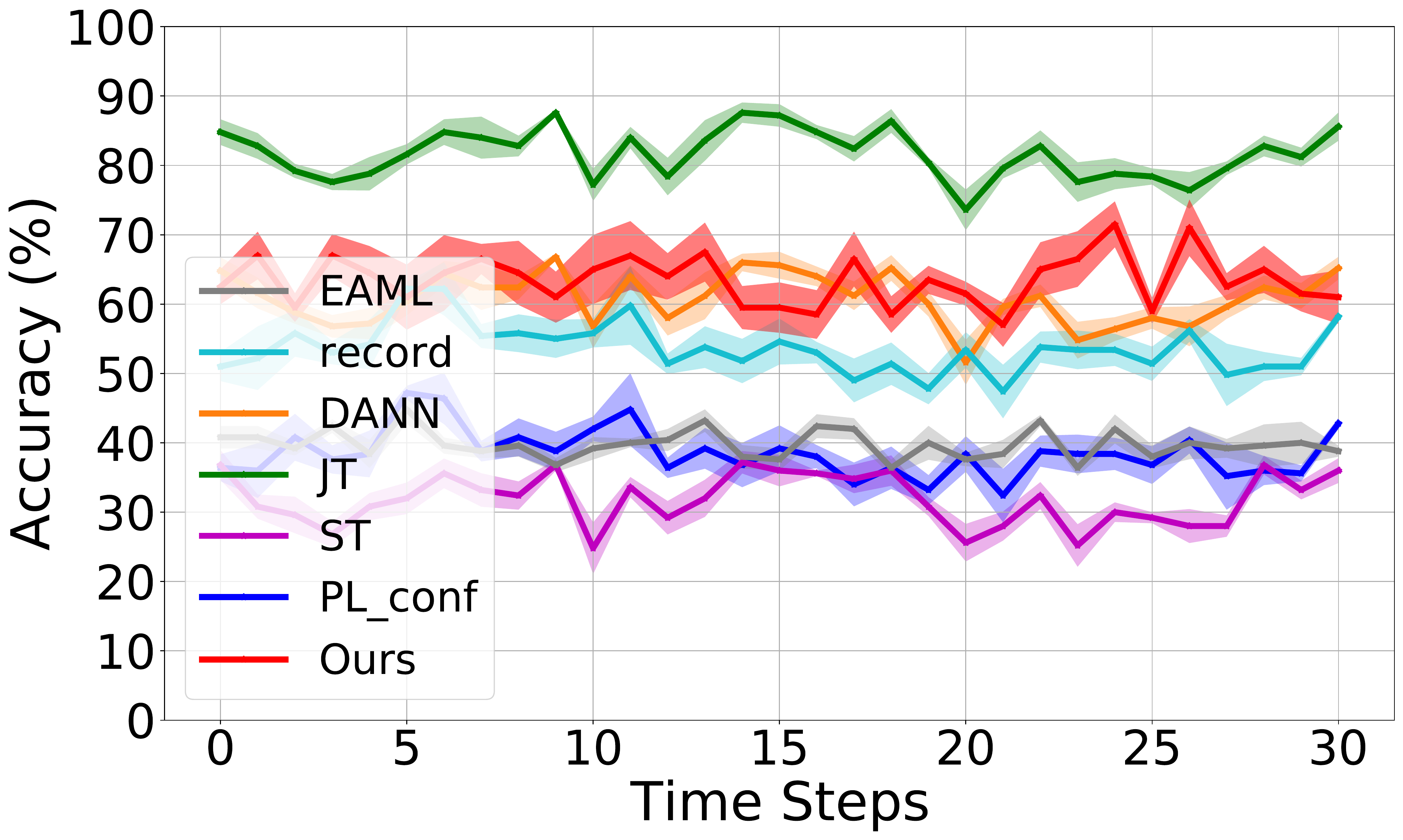}}
 \vspace{-0.3cm}
  \caption{Memorization ability through timelines. Noted that Joint Training (JT) shows the ideal performance of the SDSL setting. The closer to JT (in green curve), the better the performance.}
    \label{Fig:forget}
 \vspace{-0.5cm}
\end{figure}

\subsubsection{Effectiveness of alleviating forgetting}
To answer Q3, we validated the competence of our method on remembering previous knowledge.
Noted that JT is the upper bound of the performance since JT assumes the availability of ground-truth labels across streaming data and restores all these data for training the model.
In our experiments, we considered a sequential setting where shifted data come one after another. To validate the effectiveness of the proposed algorithm, we trained the model until the final task, and then tested the model performance on all previously seen tasks. 
All methods stop at $t= T$ after seeing all tasks.

Table~\ref{Tab:forget_competence} shows the average accuracy across all the tasks ($Acc_T$) on all the datasets. Specifically, we show the detailed results for each task along the timeline on $UG\_2C\_5D$ and Satimage datasets, which is illustrated in Figure~\ref{Fig:forget}.
The results indicate that our method achieves better performance than other baseline methods on alleviating forgetting. 
Specifically, Record takes a reply method and stores influential samples with a memory buffer on each sequential task. 
However, the restricted storing buffer determines the generalization ability and is incapable of recovering all the learned information. 
On the contrary, EAML, DANN and our method relax the storage requirements. 
But EAML uses the meta-learning strategies and is inferior to other techniques due to the lack of high-quality validation set on shifted data. 
Moreover, DANN mitigates forgetting via learning an invariant representation, which is insufficient to store distinctive knowledge about previous data. 
% Noted that our method achieves promising improvement over baselines despite of Joint training (JT) methods. 
Compared to these baseline algorithms, the performance of our method is closest to the upper bound (JT). 
The reason is that the generated high-quality pseudo labels provide exact supervisions to adjust model decision boundary, meanwhile, seek a flat minimal region to further enhance generalization ability. 
\subsubsection{Ablation study of alleviating forgetting}
\begin{figure}[t]
  \vspace{-0.3cm}
  \centering
    \subfigure[Classification performance through timelines on the Satimage data.]{
  \includegraphics[width=0.23\textwidth,height = 0.13\textheight]{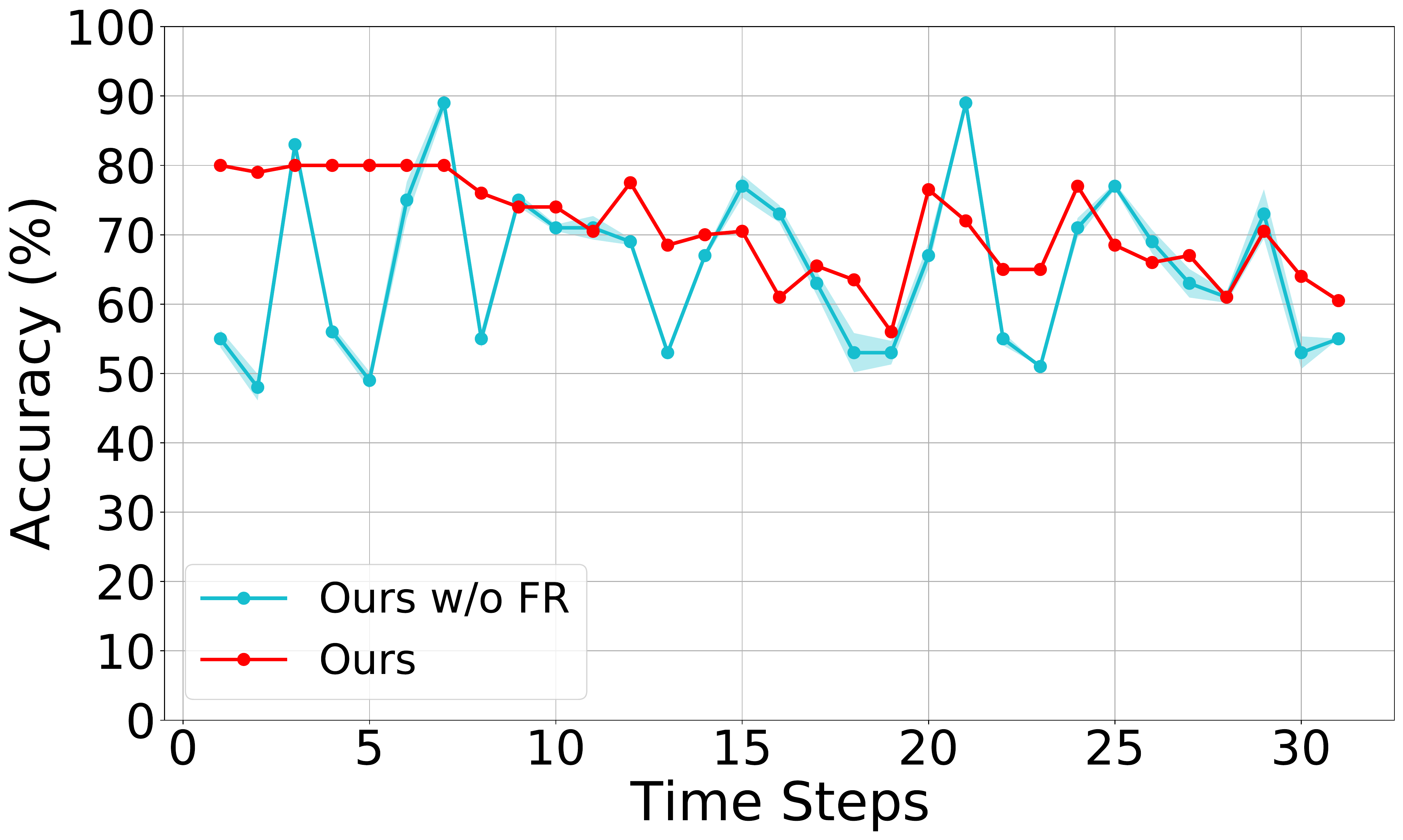}} 
  \subfigure[Memorization ability through timelines on the Satimage data.]{
  \includegraphics[width=0.23\textwidth,height = 0.13\textheight]{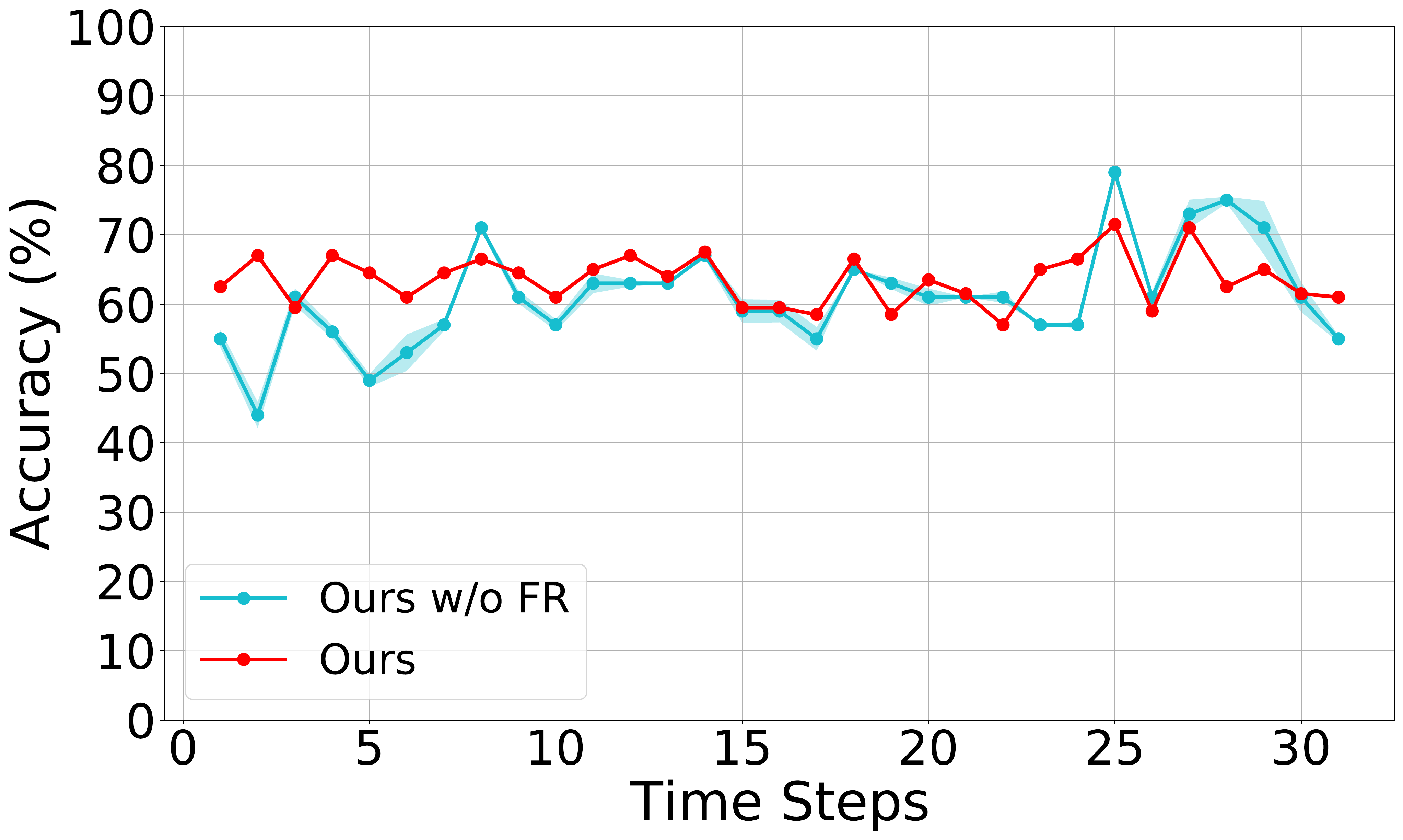}}
  % second
\vspace{-0.4cm}

\caption{Ablation study on flat region searching term.}
    % \vspace{-0.5cm}
\vspace{-0.4cm}
\label{Fig:flat_region_1}
\end{figure}
We introduce a \textbf{F}lat \textbf{R}egion (FR) constraint to better alleviate the forgetting issue. 
In the experiment, we also conduct an ablation study to investigate the contribution of the flat region constraint. 
Specifically, we compare our method with a variant that omits the FR constraint in Equation~\ref{eq:minmax}, denoted as ``Ours w/o FR''.

Figure~\ref{Fig:flat_region_1} shows our method takes merits from finding flat minimal region. The flatness of the optimal minima makes the model insensitive to parameter change. Besides, the model updates the parameters in the orthogonal direction of previous parameters space, which preserves previous knowledge when adapting to new data.

\vspace{-0.1cm}
\subsection{Empirical Validation for Flat Region Theory}
\begin{figure}[h]
\vspace{-0.5cm}
  \centering
  \subfigure[Memorization ability through timelines on the UG\_2C\_5D dataset.]{
  \includegraphics[width=0.23\textwidth,height = 0.13\textheight]{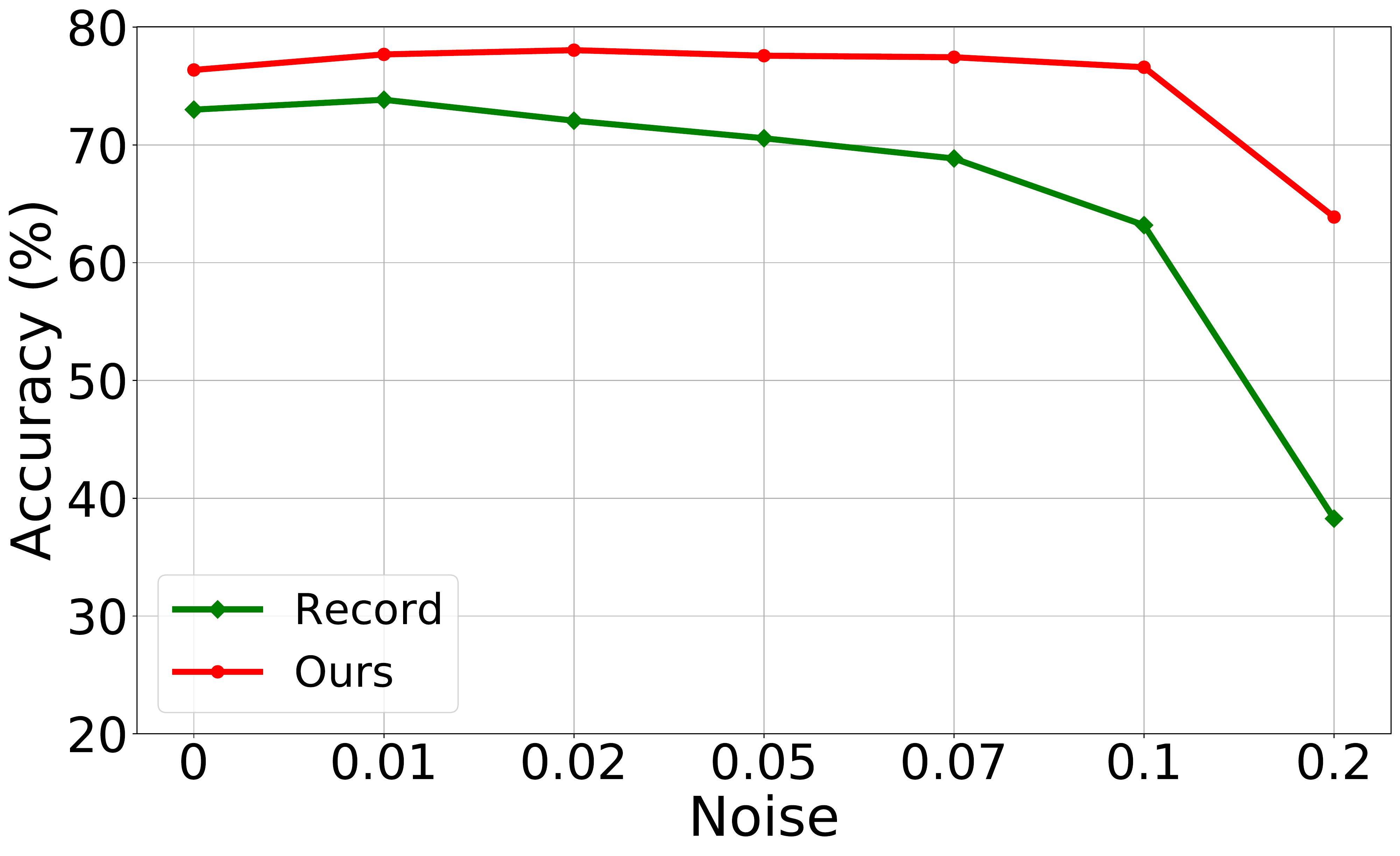}}
  % second
  \subfigure[Memorization ability through timelines on the Satimage dataset.]{
  \includegraphics[width=0.23\textwidth,height = 0.13\textheight]{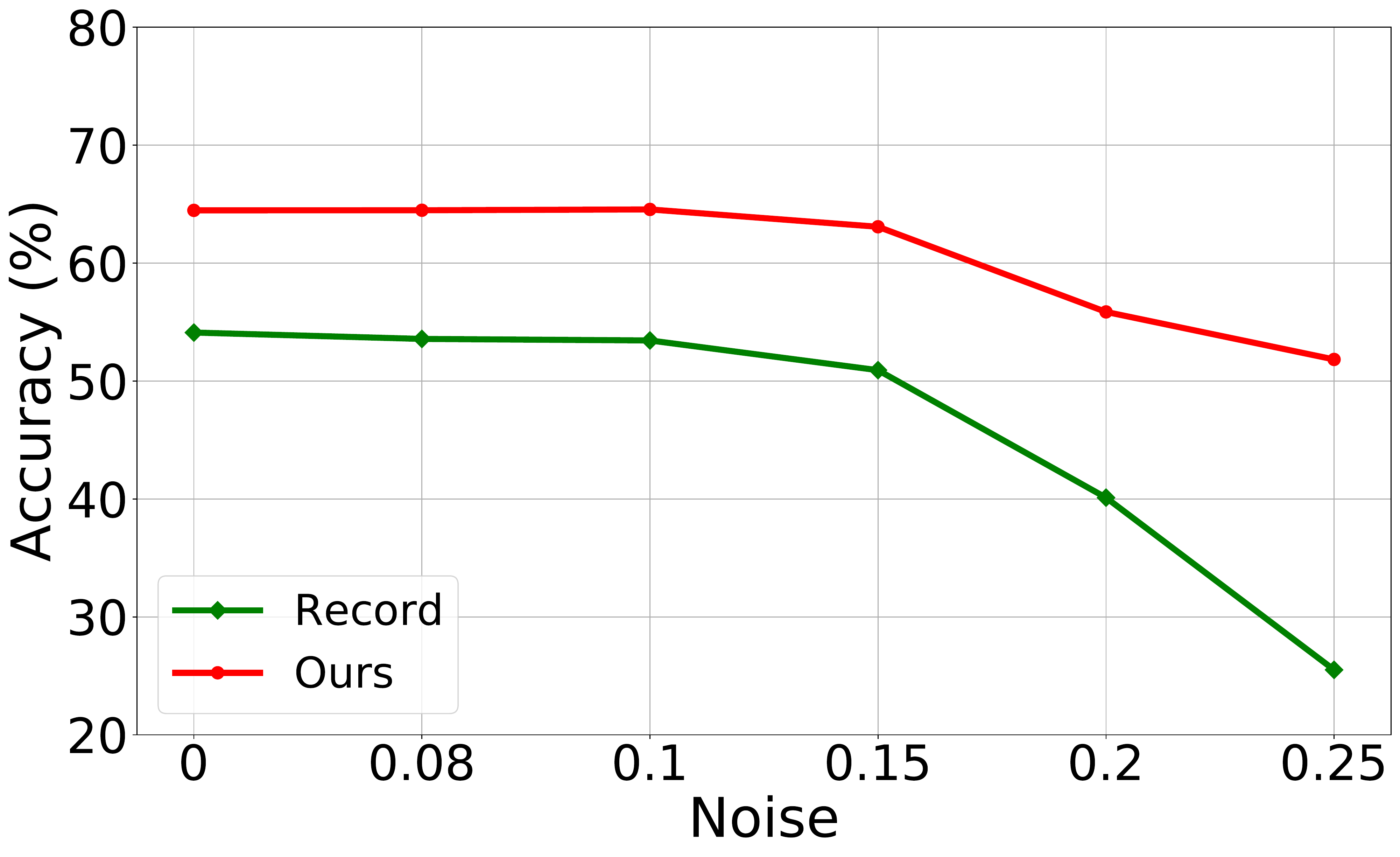}}
%   \caption{classification results w/o robust labeling component on UG\_2C\_5D and Satimage dataset through each round.}
\vspace{-0.4cm}
\caption{Flat region validation.}
\vspace{-0.2cm}
\label{Fig:Flat_region_2}
\end{figure}
\vspace{-0.2cm}

\iffalse
\begin{figure}[h]
  \centering
  \subfigure[Satimage on classification ability through time.]{
  \includegraphics[width=0.35\textwidth,height = 0.2\textheight]{samples/figure/noise_ref.pdf}}

%   \caption{Comparison of the flatness of the local optimal minima found by the Our method and baselines.}
\caption{Flat region validation.}
\label{Fig:weight_analysis}
\end{figure}
\fi

In the Appendix A.2, we introduced the theoretical analysis of the flat region concept for boosting the model generalization ability. 
The flat region enables the model insensitive to the drift, resulting in promising generalization ability.
Our question is: whether the flat region exists? If yes, can our method find a wider flat region?
% In the experiment, we also aim to provide empirical support for validating the flat region theory. 
Therefore, we conducted an empirical validation to answer the question.
Suppose the flat region exists, the deviation of the model parameters within the region will not cause significant fluctuation of the model performance. 
Therefore, in the experiment, we validated the flat region theory based on a noise-sampling method. 
Specifically, to find the local optimal $\theta^{\star}$, we measure its flatness as follows. 
We firstly sampled noise from a pre-defined region $[0,b]$, then injected the noise to the trained model parameters only in the testing phase, and reported $Acc_T$.
% We set the region upper bound b approximating to $1\%$ of current optimal parameter range. 
We changed the value of $b$, and averaged the model classification accuracy to measure its sensitiveness to noise. 
Intuitively, if the flat region exists, the model will be more robust against (less sensitive to) the injected noise. 
We compare our method with one variant that does not take flat region search (denoted as Record) in the same setting.
Due to the page limitation, we only present the result on UG\_2C\_5D and Satimage. 
%For the results on other datasets, please refer to the Appendix.
 
Figure \ref{Fig:Flat_region_2} shows that all the models have a respective flat region, but the range of the region is different. 
For example, for the UG\_2C\_5D dataset, when the upper bound $b$ changes from 0 to 0.1, the performance of our method barely changes; when $b$ is larger than 0.1, the performance begins to drop significantly. 
The observation reveals the fact that there exists a flat region $[0, 0.1]$, in which the model is insensitive to the drift. 
Similarly, we can also observe a flat region $[0, 0.15]$ for our method on the Satimage dataset.
Among the comparison, our method has the widest flat region, which means our method has the best generalization ability. 
Therefore, the empirical results validate the effectiveness of the seeking for a flat region method.
\vspace{-0.4cm}

\vspace{-0.2cm}
\section{Related Works}
Our work lies at the intersection of semi-supervised learning, continual learning (life-long learning) , and domain adaptation. Next, we provide an overview of the related research efforts and briefly discuss the connections with our work.

\noindent \textbf{Semi-Supervised Learning.} Our work is related to semi-supervised learning (SSL) ~\cite{lee2013pseudo}. 
SSL is a special case of machine learning that leverages a large amount of unlabeled data with a small portion of labeled data to enhance the learning performance.
SSL methods can be categorized into consistency-based~\cite{berthelot2019mixmatch}, temporal ensembling~\cite{laine2016temporal}, virtual adversarial training~\cite{miyato2018virtual}, pseudo labeling~\cite{lee2013pseudo}. 
Most of SSL studies are designed both for offline and ``identical and independent distribution'' (i.i.d.) data but ignore the evolving nature of unlabeled samples. 
There are some emerging works designed for streaming data in the SSL setting~ by integrating local consistency propagation on graph~\cite{zhao2022exploring,zhao2021graphsmote}. 
However, these methods assume the streaming data are i.i.d. with the labeled data which is not ideal for a realistic scenario. 
Recently, \cite{guo2020record} considers learning from streaming data with a distribution shift in a semi-supervised way. However, they still generate pseudo labels via classifiers trained on previous data and maintain a memory buffer to store pseudo labeled data sequentially. 
Differently, We improve the pseudo label generation process considering shifted data, and only replay data with short lookback.
%\textbf{Moreover, \cite{XXX} , but assuming that the labeled data is available at every time step, which is difficult to satisfy in real-world applications. }
%Comparing to the literature, our proposed method can work for real streaming shift in the SSL setting by \textcolor{red}{[one sentence to summarize]}.

\noindent \textbf{Continual Learning.} 
Our method also connects to continual learning (CL). 
CL studies the problem of learning new information throughout their lifespan, without forgetting previously learned data or tasks.
Generally speaking, there are three typical scenarios for CL: (1) class-incremental~\cite{liu2020mnemonics}, (2) task-incremental~\cite{maltoni2019continuous}, and (3) data-incremental scenario (with class label set fixed)~\cite{chrysakis2020online,ren2018tracking}. 
Our work is related to the third category, but the incoming data is shifted without any supervision. 
In contrast, we release the requirement of the availability of labeled incoming data.
%\textbf{We refer to this problem as semi supervised continual domain adaptation.}
%See~\cite{delange2021continual} for a closer look at the different problem formulations.
Different from traditional regularization-based methods that penalize any changes to previous important parameters,
%we alleviate catastrophic forgetting without necessarily replaying old data ~\cite{rolnick2019experience} nor increasing the model branch over time ~\cite{hadsell2020embracing}. 
we modify the objective function by introducing a flat region \cite{schulman2015trust,shi2021overcoming} into SDSL setting. 
Different from previous works in CL that learn the flat region with prior knowledge \cite{,schulman2015trust, shi2021overcoming}, we formulate the automated flat region identification problem as a minimax game into SDSL, which can ease the forgetting issue and adapt well to the new timeline.

\noindent \textbf{Unsupervised Domain Adaptation.} 
Our work is relevant to unsupervised domain adaptation~\cite{tzeng2017adversarial}. Unsupervised domain adaptation aims to transfer the knowledge learned from labeled source domains to the unlabeled target domain. 
Existing works mainly focus on minimizing the discrepancy between the source and target distributions for learning domain-invariant features~\cite{tzeng2017adversarial,fernando2013unsupervised}. 
% Generally, the weights of the deep architecture containing a feature encoder and a classifier are shared for both domains \cite{}.
% With the covariate shift assumption \cite{}, such invariant representations, along with the source predictor can generalize to the target domain \cite{}.
Yet, recent theoretical analysis and empirical findings suggest that distinct marginal label distributions across domains provably undermine the target generalization. 
Therefore, to minimize the distance between labeling functions, \cite{li2021learning,ren2021cross} accesses a small amount of labeled data in the target domain as extra supervision.
\cite{wang2021self} designs a contrastive re-weighting method to dynamically modify the generated labels on the target domain. Our work is enlightened by this line of works, differently, we consider a label class semantic constraints. Besides, we aim to alleviate the catastrophic forgetting problem, which is ignored by this pipeline.

\vspace{-0.3cm}
\section{Conclusion Remarks}
In this work, we provide a systematic analysis of semi-supervised drifted streaming learning with short lookback (SDSL), which is a realistic yet challenging setting without extensive study.
Then we propose a novel method that follows the 'generation-replay' pipiline.  
To generate accurate pseudo labels for incoming shifted data, we leverage supervised knowledge of previously labeled data to label overlapped data, unsupervised knowledge of new data to refine non-overlapped data, as well as structure knowledge of invariant label semantic embedding to regularize the classifier.
To achieve adaptive anti-forgetting model replay, we introduce the flat region notion and search the feasible region with a minimax game. 
Comprehensive experimental results verified our motivations and demonstrate the effectiveness of our method.  
In the future, we aim to explore
more properties of unlabeled data to further improve the robustness of SDSL setting.
\vspace{-0.4cm}

\iffalse
\begin{acks}
To Robert, for the bagels and explaining CMYK and color spaces.
\end{acks}
\fi

\normalem
\bibliographystyle{ACM-Reference-Format}
\bibliography{main}

\newpage
\appendix
\section{Appendix}
\subsection{Algorithm of Robust Pseudo-Label Generation}
\begin{algorithm}[h]
  \caption{Robust Pseudo-Label Generation}
  \label{alg:Framwork}
  %\scalebox{0.75}{
  \begin{algorithmic}[1] 
  \REQUIRE %??????????Input
    gold-label data $\mathcal{D}$,
    Pseudo-labeled set $\mathcal{\hat{D}}_{t-1}$,
    Randomly initialization on model parameter $\theta$, 
    Randomly initialization on latent feature matrix $\mathbf{H}_t$.
    \STATE Pretraining the model on golden label set $\mathcal{D}$ with feature extractor and classifier.
    \STATE Generating latent label vector $\mathbf{V}$ with classifier parameters via SVD decomposition.
    \FOR{$t$ in $T$}
    \STATE Training the model with $\mathcal{D}$ and $\mathcal{D}_{t-1}$ with $\theta$.
    \STATE Applying the model with $\theta$ on $\mathbf{X}_t$ to initialize prototypes $\mathbf{U}_t$ in Eq.\ref{eq:initialization}.
    \WHILE {Not Converged}
    \STATE Calculating $\hat{\mathbf{H}}_t$ with current $\mathbf{U}_t$ and $\mathbf{\hat{V}}$ in Eq.\ref{Eq:semantics}.
    \STATE Generating $\hat{y}_t$ in Eq.\ref{eq:label}.
    \STATE Updating prototypes $\mathbf{U}_t$ with Eq.\ref{eq:centroid} and Eq.\ref{Eq:semantics}. 
    \STATE Update $\theta$ with gradients to minimize Loss $\mathcal{L}_{full}$.
    \ENDWHILE
    \RETURN Pseudo-labeled set $\mathcal{D}_t$ and model parameter $\theta$.
    \ENDFOR
  \end{algorithmic}
  %}%% resizebox
\end{algorithm}

\subsection{Algorithm of Adaptive Anti-forgetting Model Replay}
After the robust pseudo-label generation stage, we obtain pseduo-labeled pairs $\mathcal{D}_{t}$ (also collected it with memory buffer as shown in the paper.). Then we replay the data $\mathcal{D}$ and $\mathcal{D}_{t}$, and then move to the next robust pseudo-label generation stage (t+1). Here, we represent the adaptive anti-forgetting model replay stage. 
\begin{algorithm}[h]
  \caption{daptive Anti-forgetting Model Replay}
  \label{alg:Framwork}
  %\scalebox{0.75}{
  \begin{algorithmic}[1] 
  \REQUIRE %??????????Input
    gold-label data $\mathcal{D}$,
    Pseudo-labeled set $\mathcal{\hat{D}}_t$,
    trained model parameters after time $t-1$, step size $\eta_{1}$ and $\eta_{2}$, the model parameters $\theta$ train on $t-1$
    \WHILE {Not Converged}
    \STATE update $\theta$ on $\mathcal{D}$ and $\mathcal{\hat{D}}_t$ via Eq.\ref{eq:theta}
    \STATE update $\xi $ via Eq.\ref{eq:flat} 
    \ENDWHILE
    \RETURN model parameter $\theta$.
  \end{algorithmic}
  %}%% resizebox
\end{algorithm}

\subsection{Theoretical Analysis}
We present a theoretical analysis on why the flat region can characterize the continual learning property on streaming data and why our method works. 
Without loss of generalization, we simplify the drifted stream data with $\mathcal{D}_{t-1}$ and $\mathcal{D}_{t}$, which are sampled from data distribution $Q_{t-1}$ and $Q_t$, respectively. 
Based on previous works on PAC-Bayes bound \cite{neyshabur2017exploring,deng2021flattening}, given a `prior' distribution $\mathbf{P}$ (a common assumption is zero mean, $\sigma^2$ variance Gaussian distribution) over the weights, the expected error can be bounded with probability at least 1 - $\delta$:
\begin{equation}
\begin{matrix}
%first line
&\displaystyle \min_{\Delta\theta}\mathbb{E}_{\xi}[\mathcal{L}_{\mathbf{Q}_{t-1} \cup \mathbf{Q}_t }(\theta +\Delta\theta+ \xi)]  \\
\leq 
%second line
&\displaystyle \min_{\Delta\theta \in M^c}\mathbb{E}_{\xi}[\mathcal{L}_{\mathcal{D}_{t-1} }(\theta + \Delta\theta+\xi)]
+4\sqrt{\frac{1}{n} KL(\theta+\xi||P) +ln\frac{2N}{\delta }{}}

\\
% third line
&+ \underbrace{\max_{\xi \in M}[\mathcal{L}_{\mathcal{D}_t }(\theta + \Delta\theta+ \xi)] 
-  \mathcal{L}_{\mathcal{D}_{t-1} }(\theta + \Delta \theta) }_\text{Generalization Gap} 

+\mathcal{L}_{\mathcal{D}_{t-1} }(\theta + \Delta\theta)
\end{matrix}
\end{equation}
where $\xi \in M$ and $\Delta \theta$ is updated along the orthogonal direction of previous optimal solution $\theta$ learned on previous data $\mathcal{D}_{t-1}$, i.e., $\theta \in M^c$. 
So that $\displaystyle \min_{\Delta\theta \in M^c}\mathbb{E}_{\xi}[\mathcal{L}_{\mathcal{D}_{t-1} }(\theta + \Delta\theta+\xi)]$ does not change too much compared with the previously minimized $\mathcal{L}_{\mathcal{D}_{t-1} }(\theta)$.
Similarly, the updated parameters will not increase the training loss on $\mathcal{D}_{t-1}$.
The second term depicts the Kullback Leibler (KL) divergence to the ``prior'' P \cite{neyshabur2017exploring}.
Our method exactly optimizes the worst-case of the
flatness of weight loss landscape $\displaystyle \max_{\xi \in M}\mathcal{L}_{\mathcal{D}_t }(\theta + \Delta\theta+ \xi)
-  \mathcal{L}_{\mathcal{D}_{t-1} }(\theta + \Delta \theta)$ to control the above PAC-Bayes bound, which theoretically justifies why our method works.

\subsection{Additional Implementation Details}
All experiments were conducted on the Ubuntu 18.04.5 LTS operating system, Intel(R) Core(TM) i9-10900X CPU@ 3.70GHz, and 1 way SLI RTX 3090 and 128GB of RAM, with the framework of Python 3.8.5 and PyTorch 1.8.1.

\iffalse
\section{Online Resources}
The code and data for the experiments could be downloaded at \url{https://www.dropbox.com/s/7edvhaeiypk6sl7/dataset_and_code.zip?dl=0}
\fi

\end{document}